\documentclass{article}

\usepackage{PRIMEarxiv}

\usepackage[utf8]{inputenc} 
\usepackage[T1]{fontenc}   
\usepackage{amsmath}
\usepackage{amsfonts}
\usepackage{amssymb}
\usepackage{algorithm}
\usepackage{algorithmic}
\usepackage{caption}
\usepackage{subcaption}
\usepackage{graphicx}
\usepackage{xcolor}
\usepackage{textcomp}
\usepackage{comment}
\usepackage{float}
\usepackage{pdflscape}   
\usepackage{multirow}
\usepackage{booktabs}
\usepackage{colortbl}
\usepackage{adjustbox}
\usepackage{pdflscape}
\usepackage{afterpage}
\usepackage{hyperref}
\usepackage{amsthm}
\usepackage[numbers]{natbib}

\newtheorem{proposition}{Proposition}

\title{Bi-Objective Chance-Constrained Evolutionary Optimization for Large-Scale Open-Pit Mine Scheduling under Geological Uncertainty}

\author{ \href{https://orcid.org/0000-0002-8043-994X}{\includegraphics[scale=0.06]{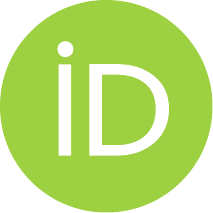}\hspace{1mm}Ishara Hewa Pathiranage} \\
	Machine Learning and Optimisation, \\School of Computer Science and \\Information Technology,\\
  Adelaide University,\\ Adelaide, Australia 
	\And
	\href{https://orcid.org/0000-0002-0036-4782}{\includegraphics[scale=0.06]{orcid.pdf}\hspace{1mm}Aneta Neumann} \\
	Machine Learning and Optimisation, \\
    School of Computer Science and\\ Information Technology,\\
  Adelaide University,\\ 
  Adelaide, Australia \\
}

\begin{document}

\maketitle

\begin{abstract}
The open-pit mine scheduling problem (OPMSP) is a complex optimization problem in long-term mine planning that involves numerous operational and geological constraints. Traditional deterministic approaches often ignore geological uncertainty, leading to suboptimal or unreliable production schedules. Chance constraints provide a framework for handling uncertainty by ensuring that probabilistic constraints are satisfied with a predefined confidence level. In this paper, we consider the OPMSP under geological grade uncertainty and propose a bi-objective chance-constrained formulation that simultaneously maximizes the expected discounted net present value and minimizes scheduling risk. Unlike traditional chance-constrained approaches, the proposed formulation does not require a predefined confidence level during optimization. Instead, it generates a set of Pareto-optimal solutions representing different trade-offs between profitability and risk within a single optimization run. To solve the resulting large-scale stochastic optimization problem, we employ multi-objective evolutionary algorithms and compare their performance against a single-objective chance-constrained evolutionary approach and a deterministic MILP benchmark. We further evaluate the contribution of the problem-specific initialization and mutation components through an ablation study. Experimental results on MineLib benchmark instances containing up to $112\,687$ blocks demonstrate that the proposed formulation effectively captures the trade-off between profitability and risk under geological uncertainty while providing greater flexibility than confidence-level-dependent approaches.
\end{abstract}


\section{Introduction}
\label{sec: introduction}
Open-pit mine scheduling is a complex and computationally challenging problem in
long-term mine planning. It involves numerous decision variables, as well as precedence and operational constraints. The main objective of this problem is to determine an optimal sequence of block extractions over multiple periods to maximize net present value (NPV), while satisfying geographical and operational constraints. Over the past years, various approaches have been proposed to address OPMSP, including integer linear programming techniques~\cite{Letelier}, heuristics~\cite{JELVEZ20161169}, and metaheuristics~\cite{ELSAYED202077,10612008, PAITHANKAR2019105507}.

Traditional OPMSP formulations assume a single deterministic block model with complete knowledge of ore grade information. However, geological uncertainty in ore grades introduces risk, potentially resulting in unreliable and suboptimal production schedules. Recent literature addresses this problem by introducing ensemble block models~\cite{PAITHANKAR2021101875,10.1145/3449726.3463135,10254112}, which contain multiple, equally probable geological realizations generated through geostatistical simulations.
Instead of relying on a single deterministic block model, ensemble-based approaches evaluate scheduling decisions across a distribution of ore bodies. This enables the development of risk-aware production schedules.

Chance constraints~\cite{Charnes} offer a framework to handle uncertainty in optimization problems by ensuring that stochastic constraints are satisfied with high probability, typically denoted by $\alpha$. This methodology has been successfully applied in domains such as mining~\cite{doi:10.1080/25726668.2021.1916170,10.1145/3449639.3459382} and logistics~\cite{10.5555/2888116.2888219, FARINA201653}.

Solving chance-constrained optimization problems for large-scale instances is computationally challenging with exact methods because of their combinatorial and nonlinear characteristics~\cite{10254112}. Evolutionary algorithms (EAs) offer a practical alternative, as they are well-suited to stochastic, nonlinear, and high-dimensional problems. In recent years, EAs have been successfully applied to a wide range of mining optimization problems~\cite{ELSAYED202077, 10612008, GILANI2020101738, KHAN2018428,9504901}.

These algorithms have proven effective for large-scale chance-constrained optimization problems. However, existing chance-constrained evolutionary approaches for OPMSP typically employ single-objective formulations with a predefined confidence level $\alpha$~\cite{10254112}.
Consequently, different reliability levels require separate optimization runs, limiting the exploration of alternative scheduling strategies under uncertainty.

Multi-objective evolutionary algorithms~(MOEAs) provide an effective framework for addressing this limitation by reformulating probabilistic constraints as additional objectives. This enables the simultaneous optimization of conflicting objectives, such as expected value and risk~\cite{AHMADI201656}. As a result, MOEAs can generate a set of Pareto-optimal solutions that represent different trade-offs between profitability and risk within a single optimization run. Although multi-objective reformulations have previously been investigated for several stochastic combinatorial optimization problems~\cite{10.1145/3638529.3654066,DBLP:conf/ppsn/NeumannN20, 10.1007/978-3-031-70055-2_8,10.1162/evco_a_00360,DBLP:conf/gecco/0001W23, 10.1145/3638529.3654067,10.1145/3795095.3805133}, their applicability to large-scale open-pit mine scheduling under geological uncertainty has not yet been investigated. Unlike these problems, OPMSP involves additional challenges, including complex mining precedence constraints, multi-period extraction scheduling, large-scale decision spaces, and geological uncertainty. To address this gap, this paper proposes a bi-objective chance-constrained formulation for large-scale OPMSP under geological uncertainty and solves it using multi-objective evolutionary algorithms.

\subsection{Related Work}

The open-pit mine scheduling problem has been widely studied due to its economic significance and computational complexity. Traditional approaches
include mixed-integer linear programming formulations~\citep{Letelier,Muñoz2018},
which guarantee optimality but become computationally expensive for large-scale,
real-world instances~\citep{ELSAYED202077}. To address these scalability limitations,
heuristic and metaheuristic approaches, particularly evolutionary algorithms,
have been widely used to explore large and complex search spaces efficiently~\citep{ELSAYED202077,10612008,9504901}. For example,~\citet{ELSAYED202077} proposed a differential evolution (DE)-based
approach that reduces problem dimensionality by considering a single-period
formulation. Their method incorporates a repair mechanism to ensure feasibility
and a local search component to improve solution quality, achieving strong
performance on instances with up to $112{,}687$ blocks. Similarly,~\citet{9504901}
formulated the stockpile blending problem as a large-scale continuous optimization
problem and introduced customized repair operators to handle constraint violations,
demonstrating improved performance over baseline methods. Furthermore,
\citet{10612008} investigated OPMSP using MOEAs, showing that NSGA-II produces higher-quality solutions than GSEMO,
particularly when combined with local search. These studies highlight the
effectiveness of EAs in handling the combinatorial structure and precedence
constraints of large-scale OPMSP instances.

Most existing approaches assume a deterministic geological model, which does not account for the inherent uncertainty in ore grade distributions. To overcome this limitation, recent studies have employed ensemble-based models that incorporate multiple geological realizations obtained through geostatistical simulation~\citep{PAITHANKAR2021101875,10.1145/3449726.3463135,10254112}.
For instance,~\citet{PAITHANKAR2021101875} proposed a stochastic optimization
framework integrating maximum flow and genetic algorithms, achieving improved
expected NPV and reduced risk. Similarly,~\citet{10.1145/3449726.3463135}
developed methods for quantifying economic uncertainty using multiple
geological interpolations. These approaches enable more realistic evaluation
of scheduling decisions under uncertainty.

Chance-constrained optimization~\citep{Charnes} provides a promising framework
for handling uncertainty by ensuring
that constraints are satisfied with a specified probability. For example,
\citet{10.1145/3449639.3459382} formulated the stockpile blending problem as a
nonlinear optimization problem with chance constraints and solved it using a
DE-based approach. Likewise,~\citet{10254112} applied a single-objective EA to
chance-constrained OPMSP by incorporating uncertainty into fitness evaluation
across multiple geological realizations. However, these approaches require a predefined confidence level $\alpha$, which limits the flexibility and exploration of trade-offs between profitability and risk.

More recently, bi-objective reformulations have also been investigated for several stochastic combinatorial optimization problems, including the stochastic knapsack problem~\cite{10.1145/3638529.3654066, 10.1007/978-3-031-70055-2_8}, submodular problems~\cite{10.1162/evco_a_00360}, and the minimum dominating set problem~\cite{DBLP:conf/gecco/0001W23}. These studies demonstrate the advantages of simultaneously optimizing expected performance and risk under uncertainty. However, they consider fundamentally different optimization problems and do not address the unique characteristics of open-pit mine scheduling, such as mining precedence constraints, multi-period extraction planning, and geological uncertainty.

Recent studies have also explored learning-based and multi-objective approaches for other complex scheduling and engineering problems. For example, \citet{WANG2026107555} proposed a machine-aligned multi-agent reinforcement learning framework for dynamic flexible job shop scheduling with stochastic job arrivals. Their method supports online scheduling by addressing non-stationarity and credit assignment in multi-agent learning. However, it considers manufacturing scheduling rather than open-pit mine scheduling under geological uncertainty and chance constraints. Similarly, \citet{11346952} formulated a bi-objective optimization problem for multi-UAV-assisted mobile edge computing, jointly minimizing energy consumption and delay, and developed an information-feedback evolutionary algorithm to improve Pareto-front convergence and diversity. These studies demonstrate the effectiveness of advanced learning and multi-objective optimization methods for complex scheduling and resource-allocation problems. Still, they do not consider geological uncertainty or chance-constrained optimization in open-pit mine scheduling.

Motivated by these developments, this paper extends the concept of bi-objective chance-constrained optimization to the large-scale OPMSP under geological uncertainty. The proposed formulation eliminates the need to specify a confidence level $\alpha$ during optimization by simultaneously maximizing the expected discounted NPV and minimizing risk. Therefore, a single optimization run generates a set of Pareto-optimal solutions, allowing decision-makers to select schedules that satisfy different reliability requirements.

\subsection{Our Contribution}

In this paper, we investigate the open-pit mine scheduling problem under geological uncertainty, where the economic value of each block is stochastic due to uncertain ore grades. Existing chance-constrained approaches rely on single-objective formulations with a predefined confidence level $\alpha$, requiring separate optimization runs for different confidence levels.

The proposed bi-objective chance-constrained formulation extends previous studies on bi-objective chance-constrained optimization to the domain of large-scale open-pit mine scheduling under geological uncertainty. In particular, it addresses the inherent challenges of OPMSP, including complex mining precedence constraints, multi-period extraction scheduling, large-scale decision spaces, and geological uncertainty. Specifically, the proposed formulation simultaneously maximizes the expected discounted net present value and minimizes scheduling risk. Furthermore, unlike traditional chance-constrained formulations, it eliminates the need to specify a confidence level during optimization. It enables the generation of a set of Pareto-optimal solutions within a single optimization run. Consequently, decision-makers can select solutions corresponding to different reliability requirements after optimization.

To efficiently solve this problem, we employ well-established multi-objective evolutionary algorithms. 
We further introduce a greedy randomized initialization procedure and a precedence-aware mutation operator tailored to OPMSP to improve the search efficiency of the proposed algorithms.
We evaluate the proposed framework using GSEMO, MOEA/D, and NSGA-II, and compare their performance against a single-objective $(1+1)$ EA baseline and a deterministic MILP benchmark. We further investigate the contribution of the proposed algorithmic components through an ablation study. Experimental results on large-scale MineLib benchmark instances containing up to $112{,}687$ blocks demonstrate that the proposed bi-objective formulation effectively captures the trade-off between profitability and risk under geological uncertainty while providing greater flexibility than traditional confidence-level-dependent approaches.

The remainder of this paper is organized as follows. Section~\ref{sec:opmspu} introduces the OPMSP under grade uncertainty and presents the mathematical formulation using chance constraints. Section~\ref{sec: bi-objective} presents the proposed bi-objective formulation and establishes its relationship to the original chance-constrained formulation. Section~\ref{sec: evolutionary_algorithms} describes the evolutionary algorithms, including the solution representation, initialization procedure, and mutation operator. Section~\ref{sec: experiments} presents the experimental setup and discusses the experimental results. Finally, Section~\ref{sec:conclusion} concludes the paper.

\section{Open Pit Mine Scheduling Problem under Grade Uncertainty}
\label{sec:opmspu}

In this section, we define the open-pit mine scheduling problem under grade
uncertainty using chance constraints. The classical open-pit mine scheduling problem considers a set of
mineable blocks $\mathcal{B}$ and a finite planning horizon
$\mathcal{T} = \{1,\dots,T\}$. Each block $b \in \mathcal{B}$ is associated with an
economic value and a set of resource requirements. A feasible solution assigns
each block to at most one period while satisfying precedence constraints between
blocks and resource capacity constraints in each period. The objective of the
classical OPMSP is to determine an extraction schedule that maximizes the total
discounted net present value.

To incorporate geological uncertainty, we extend the classical OPMSP by modeling
the economic value of each block as a random variable. Let
$x = \{x_b^t \mid b \in \mathcal{B},\, t \in \mathcal{T}\}$ denote the set of binary
decision variables, where $x_b^t = 1$ if block $b$ is mined in period $t$, and
$x_b^t = 0$ otherwise. Let $\mathcal{P}$ be the set of precedence constraints, where $(a,b) \in \mathcal{P}$
implies that block $a$ must be mined before block $b$. Each block $b$ is associated
with a stochastic economic value $p_b$, modeled as a random variable, and
discounted economic values are computed using an annual discount rate
$d \in (0,1)$.
Let $\mathcal{R}$ denote the set of resources. For each resource $r \in
\mathcal{R}$, block $b$ requires $r_b$ units, and $R_r^t$ denotes the available
capacity of resource $r$ in period $t$.

For a given schedule $x$, the total discounted net present value is a random variable
defined as
\[
p(x) = \sum_{t \in \mathcal{T}} \frac{1}{(1 + d)^t}
\sum_{b \in \mathcal{B}} p_b \, x_b^t.
\]

To ensure reliability under geological uncertainty, we incorporate a chance constraint into the discounted net present value formulation. Let $P$ denote the profit that a schedule guarantees with probability at
least $\alpha$. The chance-constrained OPMSP is formulated as
\begin{align}
\textbf{Maximize} \quad & P, \nonumber \\
\textbf{Subject to} \quad 
& \Pr\!\big(p(x) \geq P\big) \geq \alpha, \label{eq:chance-constraint}\\
& \sum_{\tau=1}^{t} x_a^\tau \geq \sum_{\tau=1}^{t} x_b^\tau,
\quad \forall (a,b) \in \mathcal{P},\; t \in \mathcal{T}, \label{eq:cons1}\\
& \sum_{t \in \mathcal{T}} x_b^t \leq 1,
\quad \forall b \in \mathcal{B}, \label{eq:cons2}\\
& \sum_{b \in \mathcal{B}} r_b \, x_b^t \leq R_r^t,
\quad \forall r \in \mathcal{R},\; t \in \mathcal{T}, \label{eq:cons3}\\
& x_b^t \in \{0,1\},
\quad \forall b \in \mathcal{B},\; t \in \mathcal{T}. \label{eq:cons4}
\end{align}
%Constraint~\eqref{eq:chance-constraint} 
The objective is to determine a schedule that maximizes the guaranteed profit $P$ while ensuring that a profit of at least $P$ is achieved with probability at least $\alpha$~(Equation~\eqref{eq:chance-constraint}). Constraints~\eqref{eq:cons1}–\eqref{eq:cons3}
enforce precedence, reserve, and resource feasibility constraints. Constraint~\eqref{eq:cons4} ensures the binary nature of the decision variables.
In this paper, we assume
$\alpha \in [1/2,1)$.

Under the assumption that the total discounted NPV is approximately normally distributed, we can reformulate the chance constraint in terms of the mean and variance of the discounted NPV. Following the approach proposed in~\cite{DBLP:conf/ppsn/NeumannXN22} for the stochastic knapsack problem, we optimize the following risk-adjusted objective:
\begin{equation}
f(x) = \mathbb{E}[p(x)] - F_\alpha \cdot \sqrt{\mathrm{Var}[p(x)]},
\label{eq:cc_p}
\end{equation}
where $\mathbb{E}[p(x)]$ and $\mathrm{Var}[p(x)]$ denote the expected value and
variance of the total discounted NPV, respectively, and $F_\alpha$ denotes the
$\alpha$-quantile of the standard normal distribution.

Since each block is extracted at most once, period NPVs are computed from
disjoint sets of blocks. 
The expected discounted NPV is given by
\begin{equation}
\mathbb{E}[p(x)] = \sum_{t \in \mathcal{T}} \mu_t =
\sum_{t \in \mathcal{T}} \frac{1}{(1 + d)^t}
\sum_{b \in \mathcal{B}} \mathbb{E}[p_b] \, x_b^t,
\label{eq:expected_npv}
\end{equation}
where $\mu_t$ denotes the expected discounted NPV in period $t$.

Following~\cite{10254112}, we compute the variance of the total discounted NPV by decomposing it into period-wise contributions. Geological
uncertainty in grade values is represented using an ensemble-based approach.
Let $\mathcal{B}^{\text{ore}} \subseteq \mathcal{B}$ denote the set of ore blocks,
and let $E = \{e_1,\dots,e_{|E|}\}$ denote the set of ensemble realizations of block grades. For each realization $e \in E$, each ore block $b$
is assigned a profit value $p_{be}$ based on the sampled grade and the economic
model.

The variance of the total discounted NPV is estimated as
\begin{equation}
\mathrm{Var}[p(x)] = \sum_{t \in \mathcal{T}} \sigma_t^2
= \sum_{t \in \mathcal{T}} \frac{1}{(1 + d)^{2t}} \, \mathrm{Var}_t(x),
\label{eq:variance_npv}
\end{equation}
where $\mathrm{Var}_t(x)$ denotes the variance of the NPV obtained in
period $t$.

The period-wise variance is computed using two components: (i) the ensemble
variance of the profit of each mined ore block, and (ii) the ensemble covariance
of profits between distinct ore blocks mined within the same period. For a given
period $t$, the variance is given by
\[
\mathrm{Var}_t(x) =
\sum_{b \in \mathcal{B}^{\text{ore}}} \sigma_b^2 \, x_b^t
+
\max\!\left(
0,
\sum_{\substack{b,b' \in \mathcal{B}^{\text{ore}} \\ b \neq b'}}
\mathrm{Cov}(b,b') \, x_b^t x_{b'}^t
\right),
\]
where the first term captures block-level profit variability, and the second term
accounts for joint variability arising from dependencies between the profits of
blocks mined in the same period.

The variance and covariance of block profits are empirically estimated from the
ensemble realizations as
\[
\sigma_b^2 =
\frac{1}{|E|} \sum_{e \in E} (p_{be} - \mu_b)^2,\]
\[\mathrm{Cov}(b,b') =
\frac{1}{|E|} \sum_{e \in E}
(p_{be} - \mu_b)(p_{b'e} - \mu_{b'}),
\]
where $\mu_b = \frac{1}{|E|} \sum_{e \in E} p_{be}$ denotes the mean ensemble
economic value of block $b$.
Following~\cite{10254112}, negative covariance terms are truncated to zero to avoid underestimating the scheduling risk due to finite-sample noise.

\section{Proposed Bi-Objective Chance-Constrained Formulation for the OPMSP}
\label{sec: bi-objective}

While the chance-constrained formulation in Equation~\eqref{eq:cc_p}
provides an effective framework for enforcing reliability at a fixed confidence level
$\alpha$, it requires $\alpha$ to be specified in advance. In practice, the choice
of $\alpha$ is problem-dependent and may vary across decision-makers. To overcome this limitation, we reformulate the chance-constrained OPMSP as a bi-objective optimization problem that simultaneously maximizes expected discounted NPV and minimizes risk. This formulation removes the need to specify a confidence level during optimization and enables the evaluation of solutions for any confidence level $\alpha \geq 1/2$ after optimization.

Accordingly, we define a bi-objective fitness function
\begin{equation}
\label{eq:2obj}
g_{2D}(x) = \left(f_1(x), f_2(x)\right)
\end{equation}
with
\begin{equation}
f_1(x) = 
\begin{cases}
\mathbb{E}[p(x)], & \text{if } v(x) = 0 \\
-v(x), & \text{otherwise}
\end{cases}
\end{equation}
and
\begin{equation}
f_2(x) = 
\begin{cases}
\sqrt{\mathrm{Var}[p(x)]}, & \text{if } v(x) = 0 \\
\mathrm{Var}[p(x)] + M \cdot v(x). & \text{otherwise} 
\end{cases}
\end{equation}
Here, $f_1(x)$ maximizes the expected discounted NPV, while $f_2(x)$ minimizes the discounted standard deviation of the NPV as a measure of scheduling risk. 
The standard deviation of the discounted NPV is adopted as the risk measure because it follows directly from the original chance-constrained formulation in Equation~\eqref{eq:cc_p}, where the expected discounted NPV is penalized by a multiple of the standard deviation. Consequently, minimizing the standard deviation preserves the theoretical relationship between the original chance-constrained and proposed bi-objective formulations. Alternative risk measures, such as Value-at-Risk (VaR), Conditional Value-at-Risk (CVaR), and downside deviation, would lead to different optimization models and are therefore left for future work.

For infeasible solutions, both objectives incorporate the total constraint violation $v(x)$, ensuring that feasible solutions are always preferred over infeasible ones. For each resource $r \in
\mathcal{R}$ and period $t$, let $\rho_r^t$ denote the amount of resource $r$ used by
solution $x$ in period $t$, and let $R_r^t$ denote the corresponding capacity
limit. The violation for resource $r$ in period $t$ is defined as
$
v_r^t(x) = \max\bigl(0,\, \rho_r^t - R_r^t\bigr)$. The violation for period $t$ is given by
$
v_t(x) = \max_{r \in \mathcal{R}} v_r^t(x)$,
and the total violation over the planning horizon is
$v(x) = \sum_{t=1}^T v_t(x)$.
This penalty formulation ensures that any feasible solution dominates any
infeasible solution in the bi-objective optimization process.

\subsection{Relationship to the Original Chance-Constrained Formulation}
\label{sec:pareto_connection}

The proposed formulation is related to recent bi-objective approaches for chance-constrained optimization that optimize the expected objective value together with its uncertainty~\cite{10.1007/978-3-031-70055-2_8}. In particular, \citet{ijcai2022p665} showed that, for chance-constrained combinatorial optimization problems, the optimal solution of the original chance-constrained formulation belongs to the Pareto-optimal set of the corresponding bi-objective formulation. We establish the corresponding relationship for the proposed chance-constrained open-pit mine scheduling problem.

For a fixed confidence level $\alpha$, let $f(x)$ denote the original chance-constrained objective defined in Equation~\eqref{eq:cc_p}.

The proposed bi-objective formulation simultaneously maximizes the expected discounted NPV and minimizes the risk, as defined in Equation~\eqref{eq:2obj}. A feasible solution $y$ dominates another feasible solution $x$ if
\[
\mathbb{E}[p(y)]\ge\mathbb{E}[p(x)],
\]
\[
\sqrt{\operatorname{Var}[p(y)]}
\le
\sqrt{\operatorname{Var}[p(x)]},
\]
and at least one of the inequalities is strict.

\begin{proposition}
\label{prop:pareto}
Let $\mathcal{F}$ denote the feasible set.
For every confidence level $\alpha>1/2$, every globally optimal solution of
\[
\max_{x\in\mathcal{F}}
\left(
\mathbb{E}[p(x)]
-
F_{\alpha}
\sqrt{\operatorname{Var}[p(x)]}
\right)
\]
is Pareto-optimal with respect to maximizing $\mathbb{E}[p(x)]$ and minimizing $\sqrt{\operatorname{Var}[p(x)]}$.
\end{proposition}

\begin{proof}
Let $x^*$ be a globally optimal solution of the scalarized single-objective. Assume that $x^*$ is not Pareto-optimal. Then there exists another feasible solution $y\in\mathcal{F}$ such that
\[
\mathbb{E}[p(y)]
\ge
\mathbb{E}[p(x^*)],
\]
\[
\sqrt{\operatorname{Var}[p(y)]}
\le
\sqrt{\operatorname{Var}[p(x^*)]},
\]
with at least one inequality being strict.
Since $F_{\alpha}>0$ for $\alpha>1/2$,
\[
-
F_{\alpha}
\sqrt{\operatorname{Var}[p(y)]}
\ge
-
F_{\alpha}
\sqrt{\operatorname{Var}[p(x^*)]}.
\]
Adding the two inequalities gives
\[
\mathbb{E}[p(y)]
-
F_{\alpha}
\sqrt{\operatorname{Var}[p(y)]}
>
\mathbb{E}[p(x^*)]
-
F_{\alpha}
\sqrt{\operatorname{Var}[p(x^*)]},
\]
which contradicts the optimality of $x^*$. Therefore, $x^*$ must be Pareto-optimal.
\end{proof}
Proposition~\ref{prop:pareto} shows that the optimal solution of the original chance-constrained formulation is contained in the Pareto-optimal set of the proposed bi-objective formulation. Therefore, if the complete Pareto-optimal set is available, we can obtain the optimal solution corresponding to any confidence level by evaluating the scalar objective over the Pareto-optimal solutions without solving the optimization problem again. In practice, however, evolutionary algorithms generate only an approximation of the Pareto-optimal set. Therefore, the recovered solution depends on the quality of the approximation. The converse does not necessarily hold. A Pareto-optimal solution is not guaranteed to be optimal for any particular confidence level, especially when the Pareto front is non-convex.

\section{Evolutionary Algorithms for Chance-Constrained OPMSP}
\label{sec: evolutionary_algorithms}

This section presents the single- and multi-objective evolutionary algorithms
used to evaluate the proposed formulations, together with the solution
representation and the domain-specific operators used for the initial feasible
solution generation and mutation.

\subsection{Integer Vector-Based Solution Representation}

We represent a scheduling solution $x$ as an integer vector
$x = (x_b)_{b \in \mathcal{B}}$, where $\mathcal{B}$ denotes the
set of blocks and $|\mathcal{B}|$ is the total number of blocks. Each decision
variable $x_b \in {-1,1,\dots,T}$ encodes the extraction period of block
$b \in \mathcal{B}$, where $x_b = -1$ denotes an unassigned block and
$x_b = t \in {1,\dots,T}$ indicates that block $b$ is scheduled in period $t$.
This representation is equivalent to the binary formulation $x_b^t$, with $x_b=t$ indicating that block $b$ is assigned to period $t$. Consequently, each block is assigned to at most one period, enabling efficient evaluation of resource and precedence constraints.

In contrast, the traditional representation uses a binary matrix of size
$|\mathcal{B}| \times T$, where each entry $(i,t)$ indicates whether block $i$
is mined in period $t$.
Compared to this matrix-based encoding, the proposed integer-vector representation reduces the dimensionality of the search space from $|\mathcal{B}| \times T$ binary variables to $|\mathcal{B}|$ integer variables and is therefore more suitable for large block models and long planning horizons.

\subsection{Single-objective Evolutionary Algorithm}

We use the single-objective (1+1)~EA~\cite{DROSTE200251} for solving the OPMSP as the baseline method. The algorithm starts from an initial feasible solution
$x = (x_1, x_2, \dots, x_{|\mathcal{B}|})$ generated using
Algorithm~\ref{algo:initial_feasible_solution}. At each iteration, an offspring
$y$ is generated using the PeriodSwapMutation operator
(Algorithm~\ref{algo:PeriodSwapMutation}) and evaluated according to
Equation~\eqref{eq:cc_p}. The offspring replaces the parent if
$f(y) \geq f(x)$.

\subsection{Multi-Objective Evolutionary Algorithms}

We evaluate the proposed bi-objective formulation using three multi-objective evolutionary algorithms (MOEAs): the Global Simple Evolutionary Multi-Objective Optimizer (GSEMO)~\cite{gsemo}, the Multi-objective Evolutionary Algorithm based on Decomposition (MOEA/D)~\cite{4358754}, and the Non-dominated Sorting Genetic Algorithm II (NSGA-II)~\cite{Deb2002AFA}.

We select these algorithms because they represent three different
multi-objective optimization paradigms. GSEMO is a simple Pareto-based
evolutionary algorithm, NSGA-II is a dominance-based approach, and
MOEA/D is a decomposition-based approach. This selection enables us to evaluate the effectiveness of the proposed formulation across different multi-objective search mechanisms.

GSEMO maintains a population $S$ of non-dominated solutions. At each iteration, a solution $x \in S$ is selected uniformly at random, and an offspring $y$ is generated. The offspring is added to the population if no existing solution in $S$ strongly dominates it. If $y$ is added to the population, all solutions in $S$ that are dominated by $y$ are removed.

MOEA/D is a decomposition-based MOEA in which each subproblem is defined by a weight vector and an aggregation function, and neighborhood relationships
among subproblems are determined based on the distances between weight vectors. MOEA/D supports several aggregation methods, including weighted sum,
Tchebycheff, and penalty boundary intersection. In this paper, we use the Tchebycheff method. The scalar subproblem is defined as
\[
g^{te}(x \mid \lambda, z^{*}) =
\max_{1 \leq i \leq m}
\left\{ \lambda_i \lvert f_i(x) - z_i^{*} \rvert \right\},
\]
where $m$ is the number of objectives, $\lambda = (\lambda_1,\ldots,\lambda_m)$
is a weight vector satisfying $\lambda_i \geq 0$ and $\sum_{i=1}^m \lambda_i = 1$,
$z^{*} = (z_1^{*}, \ldots, z_m^{*})$ denotes the ideal point, and $f_i(x)$ denotes the
$i$-th objective value of solution $x$.

NSGA-II maintains a population $S$ of size $N$ ranked by non-domination level and crowding distance. An offspring population $Q$ of size $N$ is generated using binary tournament selection and mutation. The combined population $S \cup Q$ is sorted, and the best $N$ individuals are selected for the next generation. All algorithms use the same initial feasible solution generation method and mutation operator described below.

\subsection{Initial Feasible Solution Generation}
\label{sec: Initial_Fesible_Solution}

\begin{algorithm}[!t]
\caption{Greedy–Randomized Initial Solution Generation}
\label{algo:initial_feasible_solution}
\begin{algorithmic}[1]
    \STATE \textbf{Input:} Blocks $\mathcal{B}$, horizon $T$, capacities $\{R_r^t\}$, cone values $\{c_b\}$
    \STATE \textbf{Output:} Initial schedule $x$
    \STATE Initialize $x_b \leftarrow -1$ for all $b \in \mathcal{B}$
    \FOR{$t = 1$ to $T$}
        \STATE Initialize remaining capacities $\bar{R}_r^t \leftarrow R_r^t$
        \STATE Sort unassigned blocks by decreasing $c_b$
        \FOR{each block $b$}
            \IF{$x_b = -1$ and $\mathrm{rand}() < 0.5$}
                \STATE $U \leftarrow \{b\} \cup$ unassigned predecessors of $b$
                \IF{$U$ fits within $\bar{R}_r^t$ for all $r$}
                    \STATE Assign all $u \in U$ to $t$ and update $\bar{R}_r^t$
                \ENDIF
            \ENDIF
        \ENDFOR
    \ENDFOR
    \STATE Remove trailing periods with negative discounted NPV
    \STATE \textbf{return} $x$
\end{algorithmic}
\end{algorithm}
We generate an initial feasible solution using a greedy–randomized heuristic, as outlined in Algorithm~\ref{algo:initial_feasible_solution}. The block model is represented as a directed acyclic graph to enforce precedence constraints, and each block is assigned a cone value. We compute cone values using expected profits, i.e., $c_b$ equals the sum of $\mathbb{E}[p_j]$
over block $b$ and all its predecessors~\cite{10612008}.
The algorithm iterates over periods and considers unassigned blocks in descending cone-value order, while randomized selection promotes diversification during the construction process. It assigns selected blocks along with all unassigned predecessors if remaining mining and processing capacities allow. After all periods are processed, the schedule is traversed backward to remove trailing periods with negative discounted NPV, resulting in a feasible initial solution that satisfies precedence, reserve, and resource constraints.

\subsection{Period Swap Mutation Operator}

\begin{algorithm}[!t]
\caption{PeriodSwapMutation}
\label{algo:PeriodSwapMutation}
\begin{algorithmic}[1]
    \STATE \textbf{Input:} Solution $x$, mutation probability $P_m$
    \STATE \textbf{Output:} Mutated solution $y$
    \STATE $y \gets x$
    \FORALL{$b \in \mathcal{B}$}
        \IF{$\mathrm{rand}() < P_m$}
            \STATE Define candidate periods $C_b$ based on block type (ore: earlier, waste: later, including $-1$)
            \FOR{$k=1$ to $3$}
                \STATE Sample $t' \sim \mathrm{Uniform}(C_b)$
                \IF{$t'$ satisfies precedence feasibility}
                    \STATE $y_b \gets t'$; \textbf{break}
                \ENDIF
            \ENDFOR
        \ENDIF
    \ENDFOR
    \STATE \textbf{return} $y$
\end{algorithmic}
\end{algorithm}
We use a domain-specific period-swap mutation operator during the reproduction phase (Algorithm~\ref{algo:PeriodSwapMutation}). With mutation probability $P_m$, the operator selects a block $b$ and reassigns it
to a different period or to the unassigned state. It selects candidate periods according to the block type, moving ore blocks toward earlier periods and waste blocks toward later periods. The operator samples up to three candidate periods and accepts the first reassignment that satisfies all precedence constraints, ensuring that all predecessors are scheduled no later than the new period and all successors are scheduled no earlier than the new period. If no feasible reassignment is found, the operator keeps the original assignment.

\section{Experimental Investigations}
\label{sec: experiments}

In this section, we describe the experimental setup and present results comparing the proposed bi-objective formulation with the single-objective approach for the chance-constrained OPMSP.

\subsection{Experimental Settings}
\label{sec: experimental_settings}
We conduct experiments on three benchmark instances from the MineLib library~\cite{Espinoza2013}: Newman1, Marvin, and McLaughlin Limit, with up to $112{,}687$ blocks. Table~\ref{tab:dataset} summarizes the characteristics of the three instances, including the metal types, number of periods, blocks, precedence constraints, decision variables, resources, and discount rates. Newman1 is a copper deposit characterized by a heterogeneous grade distribution. Marvin is a copper-gold deposit containing a spatially correlated high-grade zone, whereas McLaughlin Limit is a large-scale gold deposit with a heterogeneous grade distribution.

\begin{table*}[t]
\centering
\footnotesize
\caption{Summary of the MineLib benchmark instances used in the experiments~\cite{Espinoza2013}.}
\label{tab:dataset}
\begin{tabular}{l l r r r r r r}
\hline
\textbf{Instance} & \textbf{Metal Types} & \textbf{T} &
\textbf{Blocks} & \textbf{Pred.} &
\textbf{Vars.} & \textbf{Res.} & \textbf{Disc.} \\
\hline
Newman1      & Copper            & 6  & 1\,060   & 3\,922      & 6\,360      & 2 & 0.08 \\
Marvin       & Copper, Gold      & 20 & 53\,271  & 650\,631    & 1\,065\,420 & 2 & 0.10 \\
McLaughlin L.  & Gold              & 15 & 112\,687 & 3\,035\,483 & 1\,690\,305 & 1 & 0.15 \\
\hline
\end{tabular}
\end{table*}

To model geological uncertainty, we consider the economic value of each block as a random variable and generate $|E|=50$ equally probable realizations. For each block $b$, grades are sampled from a normal distribution with mean equal to the estimated grade and standard deviation set to $20\%$ of the corresponding mean value. The corresponding block profits are then computed as described in~\cite{10254112}. For a schedule $x$, the total discounted NPV, $p(x)$, is therefore a random variable. We estimate $\mathbb{E}[p(x)]$ and $\mathrm{Var}[p(x)]$ from the ensemble realizations. The variance is computed by aggregating block-level variances and empirical covariances among blocks mined within the same period.
This simplified uncertainty model enables a controlled evaluation of the proposed approach but does not explicitly capture spatial correlations, orebody continuity, or non-Gaussian grade distributions.

\begin{figure}[!t]
    \centering
    \includegraphics[width=\textwidth]{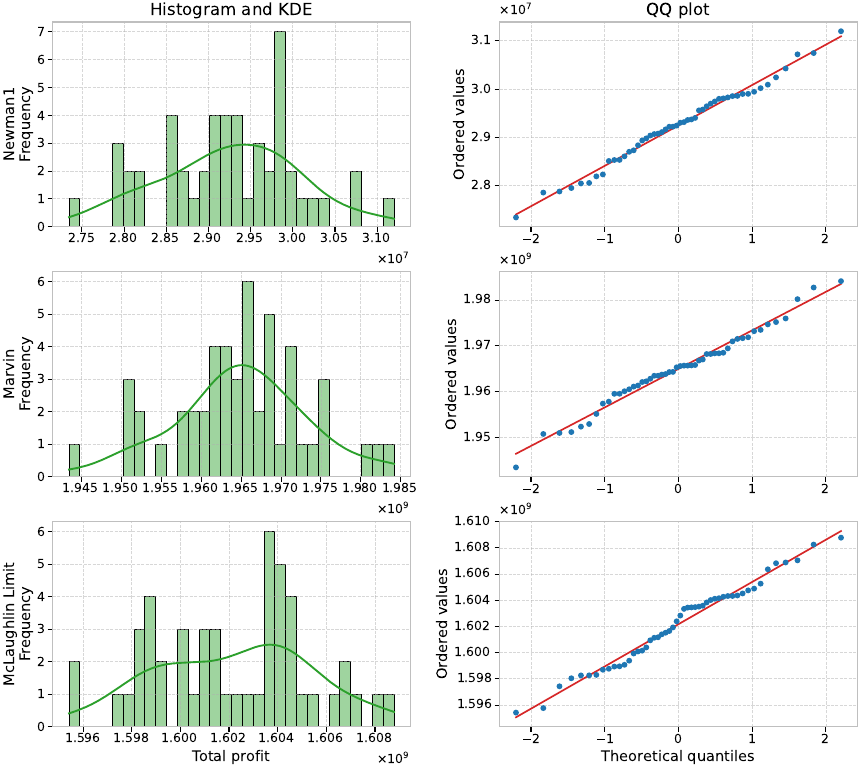}
    \caption{
    Histogram with KDE and QQ plot of the total economic value distributions obtained from the 50 geological realizations.}
    \label{fig:distribution}
\end{figure}

To validate the normality assumption underlying the proposed chance-constrained formulation, we first examine the global distribution of the total discounted NPV obtained from the $50$ geological realizations. Since the discounted NPV is obtained by aggregating the economic values of a large number of mined blocks, the Central Limit Theorem suggests that the resulting distribution is approximately normal~\cite{ross2010first}. This assumption is evaluated using histograms, QQ plots, and the Shapiro-Wilk test, as shown in Figure~\ref{fig:distribution} and Table~\ref{tab:shapiro}. In Fig.~\ref{fig:distribution}, the rows represent the Newman1, Marvin, and McLaughlin Limit instances, respectively. The left column shows the histograms with KDEs, while the right column shows the corresponding QQ plots. The histograms are approximately bell-shaped, and the QQ plot points generally
follow the reference lines. Table~\ref{tab:shapiro} reports the Shapiro-Wilk test statistic ($W$), the corresponding $p$-value, and whether the null hypothesis of normality is rejected (Rej.). For all benchmark instances, the null hypothesis of normality is not rejected at the 5\% significance level.

\begin{table}[t]
\centering
\caption{Shapiro-Wilk normality test results for the aggregate economic value distributions generated from the 50 geological realizations.}
\label{tab:shapiro}
\begin{tabular}{lccc}
\hline
\textbf{Instance} & \textbf{W} & \textbf{$p$-value} & \textbf{Rej.} \\
\hline
Newman1          & 0.9886 & 0.9077 & No \\
Marvin           & 0.9848 & 0.7614 & No \\
McLaughlin Limit & 0.9774 & 0.4464 & No \\
\hline
\end{tabular}
\end{table}

However, the chance-constrained formulation depends on the distribution of $p(x)$ for each candidate schedule rather than on the aggregate distribution. Therefore, we additionally perform a schedule-level validation. For each benchmark instance and algorithm, the representative schedule selected from each of the 30 independent runs ($\alpha=0.99$) is evaluated over all 50 geological realizations, producing 50 discounted NPV values for each schedule. A Shapiro-Wilk test is then applied to each schedule-level distribution. Table~\ref{tab:schedule-level-normality} reports the mean Shapiro-Wilk test statistic ($W$), the range of the corresponding $p$-values across the 30 schedules, and the number of schedules for which normality is rejected at the 5\% significance level (Rej.). 
Across all benchmark instances and algorithms, no schedule rejects the normality assumption, providing empirical support for the normality assumption underlying the proposed chance-constrained formulation.
Based on these empirical results, we use $F_\alpha$ as the $\alpha$-quantile of the standard normal distribution. Consistent with~\cite{10254112}, we consider
$\alpha\in\{0.60, 0.90, 0.99\}$ with
$F_\alpha\approx0.25$, $1.28$, and $2.32$, respectively.

\begin{table*}[t]
\centering
\caption{Schedule-level Shapiro-Wilk normality test results for the discounted NPV distributions. Each representative schedule obtained from an independent run was evaluated over the 50 geological realizations.} 
\label{tab:schedule-level-normality}
\begin{adjustbox}{max width=\textwidth}
\begin{tabular}{lccccccccc}
\toprule
& \multicolumn{3}{c}{\textbf{GSEMO}} & \multicolumn{3}{c}{\textbf{MOEA/D}} & \multicolumn{3}{c}{\textbf{NSGA-II}} \\
\textbf{Instance} & Mean $W$ & $p$-range & Rej. & Mean $W$ & $p$-range & Rej. & Mean $W$ & $p$-range & Rej. \\
\midrule
Newman1 & 0.9873 & 0.768 - 0.945 & 0/30 & 0.9881 & 0.749 - 0.954 & 0/30 & 0.9856 & 0.767 - 0.823 & 0/30 \\
Marvin & 0.9793 & 0.142 - 0.898 & 0/30 & 0.9808 & 0.346 - 0.802 & 0/30 & 0.9808 & 0.401 - 0.785 & 0/30 \\
McLaughlin Limit & 0.9809 & 0.365 - 0.818 & 0/30 & 0.9818 & 0.427 - 0.819 & 0/30 & 0.9826 & 0.489 - 0.783 & 0/30 \\
\bottomrule
\end{tabular}
\end{adjustbox}
\end{table*}

We evaluate the performance of the proposed bi-objective formulation using GSEMO, NSGA-II, and MOEA/D, and compare the results with those obtained by the single-objective (1+1) EA. In MOEA/D, the neighborhood
size is set to $8$, with a neighborhood selection probability of $0.9$. At most $12$ solutions are replaced in each generation, and uniformly distributed
standard weight vectors are used. In NSGA-II, the offspring population size is equal to the parent population size. We set the mutation probability to $P_m = 0.1$. The population size is set to $20$ for NSGA-II and MOEA/D, whereas GSEMO maintains an unrestricted non-dominated population during the search process. Following~\cite{10.1145/3449639.3459382}, we allocate a total budget of $10\,000$ fitness evaluations for each run. This budget provides a reasonable balance between solution quality and computational effort across all benchmark instances.

In the bi-objective formulation, a single run produces solutions for different confidence levels~$\alpha$. In contrast, for the single-objective formulation,
we divide the total evaluation budget equally across the three confidence levels. This ensures that the bi-objective and single-objective approaches are compared under the same overall computational budget as in~\cite{10.1145/3638529.3654066,DBLP:conf/gecco/0001W23}.
We evaluate the performance using the mean and standard deviation
of chance-constrained NPV over 30 independent runs.
The runs are matched by random seed across the compared algorithms to enable paired statistical comparisons.

For each confidence level $\alpha$, the solution that maximizes the original chance-constrained objective in Equation~\eqref{eq:cc_p} is selected from the obtained Pareto set. The selected solution is then used to report the chance-constrained NPV values in Table~\ref{tab:cc-npv}.

We perform statistical comparisons using the paired Wilcoxon signed-rank test at the 95\% confidence level~\cite{Wilcoxon}. Holm's procedure is used to adjust the resulting $p$-values for multiple pairwise algorithm comparisons within each benchmark instance and confidence level~\cite{Demsar2006,Benavoli2016,Carrasco2020}. The matched-pairs rank-biserial correlation is reported as the effect size~\cite{Kerby2014}.

We further evaluate the quality of the obtained Pareto-front approximations using the hypervolume (HV), inverted generational distance (IGD), generational distance (GD), and additive $\epsilon$ indicators~\cite{Zitzler2003}. Since these indicators are defined for minimization problems, the expected NPV objective is negated, while the standard deviation objective remains unchanged. Since the true Pareto front is unknown for the considered large-scale benchmark instances, an approximate reference Pareto front is constructed by merging the final non-dominated solutions obtained from all independent runs and all compared algorithms. The quality indicators are then computed with respect to this approximate reference front~\cite{Zitzler2003,DBLP:conf/gecco/DerbelLZVAT18}. The objective values are normalized to $[0,1]$ using the minimum and maximum objective values observed in the combined solution set for the corresponding instance. The HV is computed using the normalized reference point $(1.1,1.1)$, which is chosen slightly outside the normalized objective range so that all Pareto solutions weakly dominate the reference point. The IGD, GD, and additive $\epsilon$ indicators are computed with respect to the normalized reference front. The reported indicator values correspond to the mean and standard deviation over 30 independent runs.

As an additional benchmark, we implement a deterministic mixed-integer linear programming (MILP) formulation of the OPMSP and solve it using Gurobi~11.0.0. The formulation maximizes the discounted NPV subject to block-extraction, resource-capacity, and precedence constraints. Gurobi is executed on the same high-performance computing system using 32 CPU threads and 128~GB of RAM with a relative MIP optimality gap of $1\%$. The time limits are set to $90$, $2\,000$, and $15\,000$ seconds for Newman1, Marvin, and McLaughlin Limit, respectively. These limits were selected based on the observed mean wall-clock runtime of the evolutionary algorithms on each benchmark instance. This provides Gurobi with a computational budget comparable to that of the evolutionary algorithms, allowing a fair comparison.

In addition, we perform an ablation study to evaluate the contribution of the proposed initialization and mutation operators. 
We consider two ablated variants. \textit{Init Ablation} replaces the proposed greedy-randomized initialization with a uniformly random initialization while preserving the same feasibility checks and construction procedure. \textit{Mutation Ablation} replaces the proposed period-swap mutation with a standard random integer mutation that reassigns selected blocks to mining periods without using ore and waste information. In both variants, all other components and algorithmic parameters remain unchanged.

\subsection{Experimental Results}
\label{sec: experiments_results}
Next, we evaluate the effectiveness of the bi-objective formulation for the chance-constrained OPMSP and compare it with the baseline single-objective approach.

\begin{table}
\centering
\caption{Mean, standard deviation, and statistical comparison of chance-constrained discounted NPV (\$M).}
\label{tab:cc-npv}
\begin{adjustbox}{max width=\textwidth}
\begin{tabular}{llrrcrrcrrcrrc}
\toprule
\multirow{2}{*}{Instance} & \multirow{2}{*}{$\alpha$} & \multicolumn{3}{c}{\textbf{(1+1)~EA\textsuperscript{(1)}}} & \multicolumn{3}{c}{\textbf{GSEMO\textsuperscript{(2)}}} & \multicolumn{3}{c}{\textbf{MOEA/D\textsuperscript{(3)}}} & \multicolumn{3}{c}{\textbf{NSGA-II\textsuperscript{(4)}}} \\
 & & Mean & Std & Stat & Mean & Std & Stat & Mean & Std & Stat & Mean & Std & Stat \\
\midrule
\multirow{3}{*}{Newman1} & 0.60 & 23.66 & 0.09 & 2\textsuperscript{(+)}3\textsuperscript{(-)}4\textsuperscript{(+)} & 23.60 & 0.11 & 1\textsuperscript{(-)}3\textsuperscript{(-)}4\textsuperscript{(+)} & \textbf{23.74} & 0.07 & 1\textsuperscript{(+)}2\textsuperscript{(+)}4\textsuperscript{(+)} & 23.34 & \cellcolor{gray!25}0.01 & 1\textsuperscript{(-)}2\textsuperscript{(-)}3\textsuperscript{(-)} \\
 & 0.90 & 22.99 & 0.10 & 2\textsuperscript{(*)}3\textsuperscript{(-)}4\textsuperscript{(+)} & 22.93 & 0.11 & 1\textsuperscript{(*)}3\textsuperscript{(-)}4\textsuperscript{(+)} & \textbf{23.07} & 0.07 & 1\textsuperscript{(+)}2\textsuperscript{(+)}4\textsuperscript{(+)} & 22.67 & \cellcolor{gray!25}0.01 & 1\textsuperscript{(-)}2\textsuperscript{(-)}3\textsuperscript{(-)} \\
 & 0.99 & 22.33 & 0.09 & 2\textsuperscript{(+)}3\textsuperscript{(-)}4\textsuperscript{(+)} & 22.26 & 0.11 & 1\textsuperscript{(-)}3\textsuperscript{(-)}4\textsuperscript{(+)} & \textbf{22.39} & 0.07 & 1\textsuperscript{(+)}2\textsuperscript{(+)}4\textsuperscript{(+)} & 22.00 & \cellcolor{gray!25}0.01 & 1\textsuperscript{(-)}2\textsuperscript{(-)}3\textsuperscript{(-)} \\
\midrule
\multirow{3}{*}{Marvin} & 0.60 & 811.82 & 6.52 & 2\textsuperscript{(*)}3\textsuperscript{(-)}4\textsuperscript{(-)} & 810.48 & 5.05 & 1\textsuperscript{(*)}3\textsuperscript{(-)}4\textsuperscript{(-)} & \textbf{821.60} & 2.39 & 1\textsuperscript{(+)}2\textsuperscript{(+)}4\textsuperscript{(+)} & 816.40 & \cellcolor{gray!25}2.17 & 1\textsuperscript{(+)}2\textsuperscript{(+)}3\textsuperscript{(-)} \\
 & 0.90 & 807.93 & 7.91 & 2\textsuperscript{(*)}3\textsuperscript{(-)}4\textsuperscript{(-)} & 805.88 & 5.04 & 1\textsuperscript{(*)}3\textsuperscript{(-)}4\textsuperscript{(-)} & \textbf{816.97} & 2.37 & 1\textsuperscript{(+)}2\textsuperscript{(+)}4\textsuperscript{(+)} & 811.76 & \cellcolor{gray!25}2.16 & 1\textsuperscript{(+)}2\textsuperscript{(+)}3\textsuperscript{(-)} \\
 & 0.99 & 802.37 & 5.43 & 2\textsuperscript{(*)}3\textsuperscript{(-)}4\textsuperscript{(-)} & 801.22 & 5.03 & 1\textsuperscript{(*)}3\textsuperscript{(-)}4\textsuperscript{(-)} & \textbf{812.28} & 2.36 & 1\textsuperscript{(+)}2\textsuperscript{(+)}4\textsuperscript{(+)} & 807.08 & \cellcolor{gray!25}2.15 & 1\textsuperscript{(+)}2\textsuperscript{(+)}3\textsuperscript{(-)} \\
\midrule
\multirow{3}{*}{McLaughlin Limit} & 0.60 & 952.72 & 1.15 & 2\textsuperscript{(*)}3\textsuperscript{(-)}4\textsuperscript{(*)} & 952.44 & 1.18 & 1\textsuperscript{(*)}3\textsuperscript{(-)}4\textsuperscript{(-)} & \textbf{954.93} & 0.55 & 1\textsuperscript{(+)}2\textsuperscript{(+)}4\textsuperscript{(+)} & 953.20 & \cellcolor{gray!25}0.43 & 1\textsuperscript{(*)}2\textsuperscript{(+)}3\textsuperscript{(-)} \\
 & 0.90 & 950.07 & 1.41 & 2\textsuperscript{(*)}3\textsuperscript{(-)}4\textsuperscript{(*)} & 949.88 & 1.18 & 1\textsuperscript{(*)}3\textsuperscript{(-)}4\textsuperscript{(-)} & \textbf{952.36} & 0.55 & 1\textsuperscript{(+)}2\textsuperscript{(+)}4\textsuperscript{(+)} & 950.63 & \cellcolor{gray!25}0.43 & 1\textsuperscript{(*)}2\textsuperscript{(+)}3\textsuperscript{(-)} \\
 & 0.99 & 947.09 & 1.64 & 2\textsuperscript{(*)}3\textsuperscript{(-)}4\textsuperscript{(-)} & 947.29 & 1.18 & 1\textsuperscript{(*)}3\textsuperscript{(-)}4\textsuperscript{(-)} & \textbf{949.77} & 0.55 & 1\textsuperscript{(+)}2\textsuperscript{(+)}4\textsuperscript{(+)} & 948.04 & \cellcolor{gray!25}0.43 & 1\textsuperscript{(+)}2\textsuperscript{(+)}3\textsuperscript{(-)} \\
\bottomrule
\end{tabular}
\end{adjustbox}
\end{table}

Table~\ref{tab:cc-npv} presents the mean, standard deviation, and statistical comparisons of the risk-adjusted discounted objective value
(Eq.~\eqref{eq:cc_p}) obtained by four evolutionary algorithms, including three multi-objective evolutionary algorithms and one single-objective baseline across three benchmark instances and confidence levels $\alpha \in \{0.60, 0.90, 0.99\}$. The \textit{Instance} column shows the mining instance, and $\alpha$ denotes the confidence level. The remaining columns present the mean and standard deviation values, along with statistical comparisons for each algorithm, where $X^{(+)}$ indicates that the algorithm in the column outperforms algorithm $X$, $X^{(-)}$ indicates the opposite, and $X^{(*)}$ denotes no statistically significant difference. Bold values indicate the highest mean NPV per instance and confidence level, while gray-shaded cells highlight the lowest standard deviation. The detailed pairwise comparison results, including the adjusted $p$-values and effect sizes, are provided in Table~\ref{tab:wilcoxon-holm-effect_cc_1} in Appendix~\ref{app:effect_size}.

Across all benchmark instances, the discounted chance-constrained NPV decreases as the confidence level $\alpha$ increases. This behavior is expected in chance-constrained formulations, as stricter reliability requirements reduce the feasibility of high-profit schedules. Overall, MOEA/D achieves the highest mean chance-constrained NPV across most benchmark instances and confidence levels, often yielding statistically significant improvements over the competing algorithms. The advantage of MOEA/D is more pronounced for the larger instances and stricter confidence levels, indicating strong scalability under uncertainty and effective exploration of high-value extraction schedules.

NSGA-II consistently exhibits lower variance, indicating more stable convergence behavior and reduced variability across runs. However, this stability is often associated with lower expected NPV values, particularly for the Newman1 instance. In contrast, the single-objective $(1+1)$~EA and GSEMO achieve competitive performance but generally provide weaker trade-offs between profitability and risk compared to MOEA/D.

\begin{figure}
    \centering
    \includegraphics[width=\textwidth]{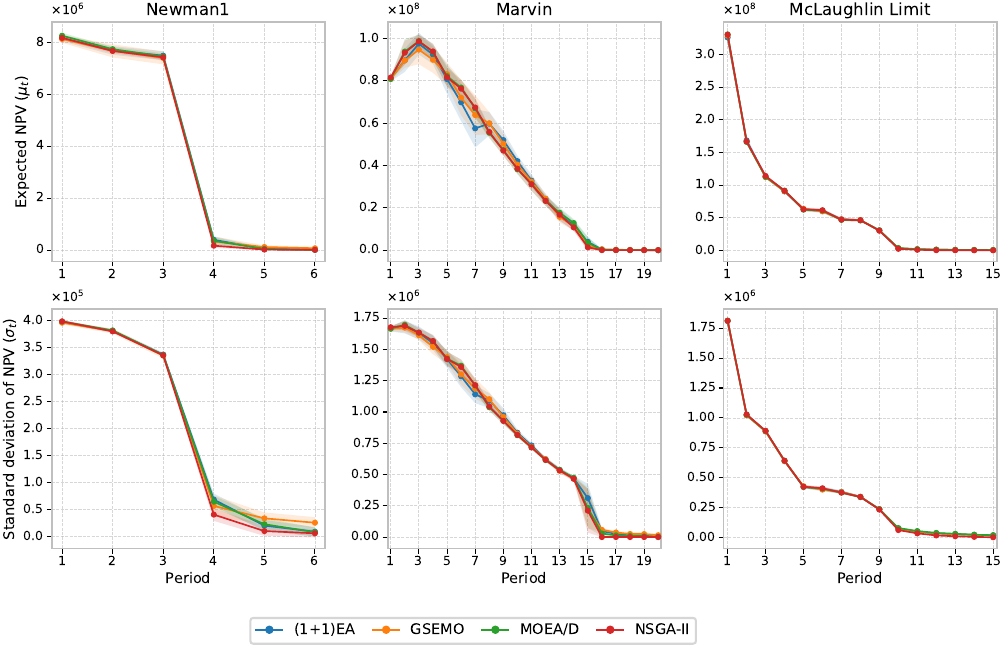}
    \caption{Yearly discounted expected NPV~($\mu_t$) (left) and discounted standard deviation~($\sigma_t$) (right) for confidence level $\alpha = 0.99$.}
\label{fig:npv_t}
\end{figure}

Furthermore, we analyze the temporal behavior of the yearly discounted expected NPV, $\mu_t$, and its discounted standard deviation, $\sigma_t$, over the mine life. Since $\mu_t$ and $\sigma_t$ are computed from ensemble realizations of the selected schedules, the overall temporal trends are similar across confidence levels. Therefore, we present representative results for $\alpha = 0.99$. Fig.~\ref{fig:npv_t} illustrates $\mu_t$, in the top row and $\sigma_t$, in the bottom row for each period. The columns represent the Newman1, Marvin, and McLaughlin Limit instances, respectively. The $x$-axis of each plot represents the scheduling period.

MOEA/D generally achieves higher discounted expected NPV values during the early production periods, indicating a stronger preference for extracting high-value blocks earlier in the schedule. This behavior is particularly evident for the Newman1 and Marvin instances. In contrast, NSGA-II produces schedules with lower temporal variability, reflected by smaller values of $\sigma_t$, resulting in more stable extraction plans.

The single-objective $(1+1)$~EA and GSEMO exhibit larger fluctuations in $\sigma_t$, particularly in later periods. Overall, the results suggest that the bi-objective formulation produces schedules that achieve a better balance between profitability and risk under geological uncertainty. These observations are consistent with the deterministic scheduling behavior reported by~\citet{Espinoza2013}.

\begin{figure}[!t]
    \centering
    \includegraphics[width=\textwidth]{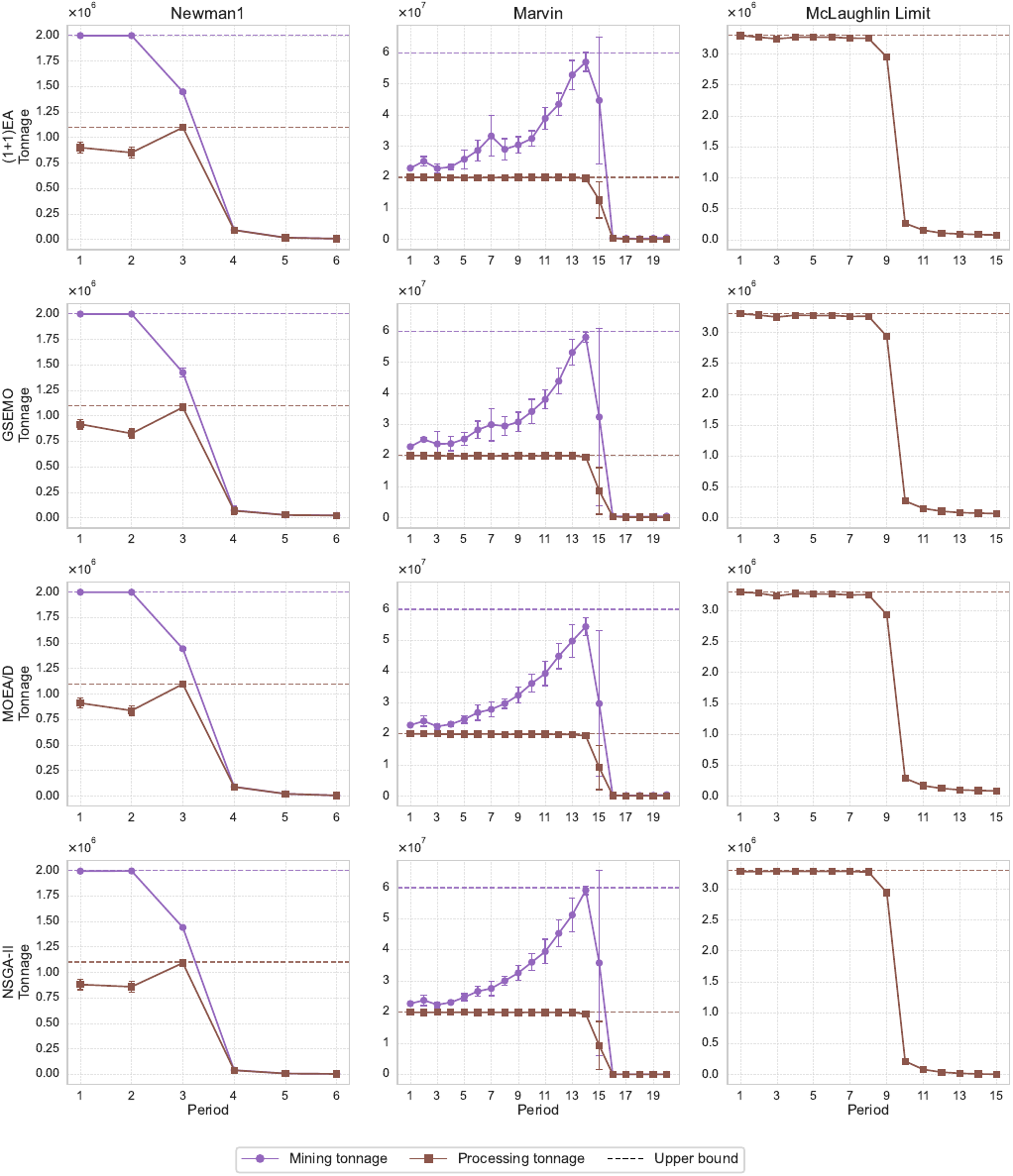}
    \caption{Mean and standard deviation of mining and processing tonnage across periods for confidence level $\alpha = 0.99$. 
    }
    \label{fig:tonnage}
\end{figure}

Fig.~\ref{fig:tonnage} compares the mining and processing tonnage across periods for $\alpha = 0.99$. The rows represent the (1+1)EA, GSEMO, MOEA/D, and NSGA-II algorithms, while the columns represent the Newman1, Marvin, and McLaughlin Limit instances, respectively. The $x$-axis represents the scheduling period, and the $y$-axis shows the mean tonnage. The dashed lines indicate the corresponding resource upper bounds. The bi-objective algorithms, particularly MOEA/D and NSGA-II, produce smoother production schedules with lower variability across the planning horizon. In contrast, the single-objective approaches exhibit more aggressive extraction patterns and larger fluctuations in production tonnage. These results suggest that explicitly incorporating risk as an optimization objective leads to more balanced production schedules while maintaining competitive economic performance. Such schedules are desirable in practical mine planning, as they promote operational consistency and reduce sensitivity to geological uncertainty.

\begin{figure}
    \centering
    \includegraphics[width=\textwidth]{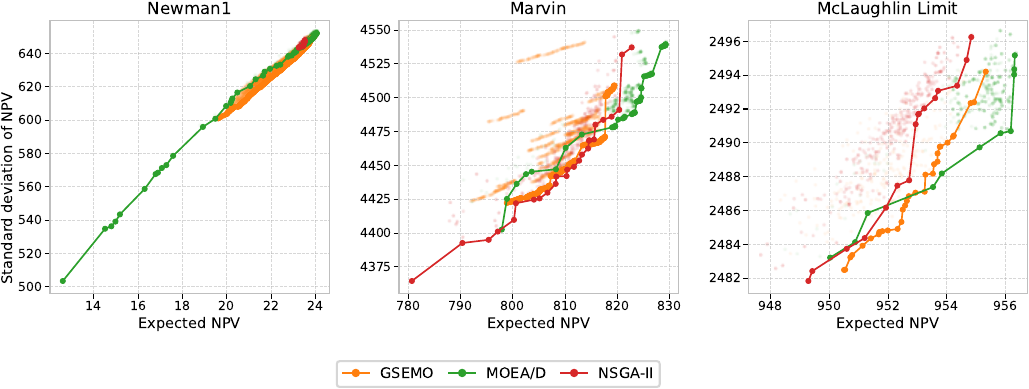}
    \caption{
Approximate Pareto fronts obtained by the three multi-objective algorithms over $30$ independent runs. Points represent the combined non-dominated solutions, and lines show the corresponding approximate Pareto fronts. The x axis shows the expected NPV (in millions; higher is better), and the y axis shows the standard deviation of NPV (in thousands; lower is better).
}
\label{fig:pareto_fronts}
\end{figure}

Next, we analyze the quality of the Pareto-front approximations produced by the three multi-objective algorithms. Fig.~\ref{fig:pareto_fronts} presents the aggregated Pareto fronts obtained across 30 independent runs. The columns represent the Newman1, Marvin, and McLaughlin Limit instances, respectively. The $x$-axis represents expected NPV, and the $y$-axis represents its standard deviation as a measure of risk. The results show differences in the coverage and distribution of the trade-offs between expected NPV and risk across the benchmark instances. For Newman1, MOEA/D provides the broadest coverage of the trade-off region, including both low-risk solutions and solutions with high expected NPV. GSEMO mainly covers the higher-NPV region, while NSGA-II produces a more concentrated approximation with a smaller spread. For Marvin, the Pareto-front approximations overlap over a substantial part of the trade-off region. MOEA/D reaches the highest-NPV solutions, whereas NSGA-II extends further toward the lower-risk region. For McLaughlin Limit, MOEA/D again reaches the highest-NPV region, while NSGA-II provides greater coverage of the lower-risk region. GSEMO mainly provides solutions within the intermediate trade-off region.

\begin{table*}[!t]
\centering
\caption{Mean, standard deviation, and statistical comparison of performance indicators over 30 paired runs. Higher values indicate better performance for HV, whereas lower values indicate better performance for IGD, GD, and $\epsilon$. 
}
\label{tab:all_indicators}
\begin{adjustbox}{max width=\textwidth}
\begin{tabular}{llrrcrrcrrc}
\toprule
\multirow{2}{*}{Instance} & \multirow{2}{*}{Indicator} & \multicolumn{3}{c}{GSEMO\textsuperscript{(1)}} & \multicolumn{3}{c}{MOEA/D\textsuperscript{(2)}} & \multicolumn{3}{c}{NSGA-II\textsuperscript{(3)}} \\
\cmidrule(lr){3-5}\cmidrule(lr){6-8}\cmidrule(lr){9-11}
 & & Mean & Std & Stat & Mean & Std & Stat & Mean & Std & Stat \\
\midrule
\multirow{4}{*}{Newman1} & HV & 0.334 & 0.043 & 2\textsuperscript{(*)}3\textsuperscript{(+)} & \textbf{0.348} & 0.067 & 1\textsuperscript{(*)}3\textsuperscript{(+)} & 0.158 & 0.003 & 1\textsuperscript{(-)}2\textsuperscript{(-)} \\
 & IGD & \textbf{0.044} & 0.016 & 2\textsuperscript{(+)}3\textsuperscript{(+)} & 0.089 & 0.014 & 1\textsuperscript{(-)}3\textsuperscript{(+)} & 0.164 & 0.003 & 1\textsuperscript{(-)}2\textsuperscript{(-)} \\
 & GD & 0.011 & 0.004 & 2\textsuperscript{(-)}3\textsuperscript{(+)} & \textbf{0.006} & 0.002 & 1\textsuperscript{(+)}3\textsuperscript{(+)} & 0.017 & 0.001 & 1\textsuperscript{(-)}2\textsuperscript{(-)} \\
 & $\epsilon$ & 0.765 & 0.050 & 2\textsuperscript{(-)}3\textsuperscript{(+)} & \textbf{0.616} & 0.205 & 1\textsuperscript{(+)}3\textsuperscript{(+)} & 0.949 & 0.003 & 1\textsuperscript{(-)}2\textsuperscript{(-)} \\
\midrule
\multirow{4}{*}{Marvin} & HV & 0.374 & 0.092 & 2\textsuperscript{(-)}3\textsuperscript{(-)} & 0.457 & 0.050 & 1\textsuperscript{(+)}3\textsuperscript{(-)} & \textbf{0.568} & 0.044 & 1\textsuperscript{(+)}2\textsuperscript{(+)} \\
 & IGD & 0.235 & 0.088 & 2\textsuperscript{(-)}3\textsuperscript{(-)} & 0.186 & 0.043 & 1\textsuperscript{(+)}3\textsuperscript{(-)} & \textbf{0.107} & 0.016 & 1\textsuperscript{(+)}2\textsuperscript{(+)} \\
 & GD & 0.102 & 0.096 & 2\textsuperscript{(-)}3\textsuperscript{(-)} & \textbf{0.046} & 0.024 & 1\textsuperscript{(+)}3\textsuperscript{(+)} & 0.063 & 0.014 & 1\textsuperscript{(+)}2\textsuperscript{(-)} \\
 & $\epsilon$ & 0.601 & 0.126 & 2\textsuperscript{(-)}3\textsuperscript{(-)} & 0.485 & 0.092 & 1\textsuperscript{(+)}3\textsuperscript{(-)} & \textbf{0.335} & 0.084 & 1\textsuperscript{(+)}2\textsuperscript{(+)} \\
\midrule
\multirow{4}{*}{McLaughlin Limit} & HV & 0.428 & 0.110 & 2\textsuperscript{(*)}3\textsuperscript{(*)} & \textbf{0.472} & 0.094 & 1\textsuperscript{(*)}3\textsuperscript{(*)} & 0.426 & 0.047 & 1\textsuperscript{(*)}2\textsuperscript{(*)} \\
 & IGD & 0.297 & 0.095 & 2\textsuperscript{(*)}3\textsuperscript{(*)} & 0.299 & 0.056 & 1\textsuperscript{(*)}3\textsuperscript{(*)} & \textbf{0.273} & 0.049 & 1\textsuperscript{(*)}2\textsuperscript{(*)} \\
 & GD & 0.172 & 0.092 & 2\textsuperscript{(*)}3\textsuperscript{(+)} & \textbf{0.131} & 0.045 & 1\textsuperscript{(*)}3\textsuperscript{(+)} & 0.271 & 0.024 & 1\textsuperscript{(-)}2\textsuperscript{(-)} \\
 & $\epsilon$ & 0.523 & 0.131 & 2\textsuperscript{(-)}3\textsuperscript{(-)} & 0.448 & 0.082 & 1\textsuperscript{(+)}3\textsuperscript{(-)} & \textbf{0.386} & 0.058 & 1\textsuperscript{(+)}2\textsuperscript{(+)} \\
\bottomrule
\end{tabular}
\end{adjustbox}
\end{table*}

While Table~\ref{tab:cc-npv} compares the algorithms using chance-constrained NPV values for specific confidence levels, the bi-objective formulation generates entire Pareto fronts. Therefore, it is also important to evaluate the quality of these Pareto-front approximations in terms of convergence and diversity. To this end, we use the HV, IGD, GD, and additive $\epsilon$ indicators.

Table~\ref{tab:all_indicators} reports the mean and standard deviation of the HV, IGD, GD, and additive $\epsilon$ indicators over 30 independent runs. We perform statistical comparisons using the same statistical tests as in Table~\ref{tab:cc-npv}. The detailed pairwise comparison results, including the adjusted $p$-values and effect sizes, are provided in Table~\ref{tab:indicator-wilcoxon-holm_pe_1} in Appendix~\ref{app:effect_size}. The results support the observations from Figure~\ref{fig:pareto_fronts}. MOEA/D achieves the best GD values across all benchmark instances, indicating best convergence toward the reference Pareto front. Furthermore, MOEA/D obtains the highest HV values for Newman1 and McLaughlin Limit, demonstrating strong coverage of the trade-off surface. These observations are consistent with the Pareto-front coverage and higher-NPV solutions observed in Figure~\ref{fig:pareto_fronts}. 
For Marvin, NSGA-II achieves the highest HV value together with the best IGD and $\epsilon$ values, indicating a well-distributed and diverse approximation of the Pareto front. This observation is consistent with the substantial overlap between the Pareto fronts shown in Figure~\ref{fig:pareto_fronts}. Overall, the indicator results suggest that MOEA/D provides consistently strong convergence and competitive diversity across benchmark instances, whereas NSGA-II achieves particularly strong diversity performance on Marvin.

\subsubsection{Comparison with Deterministic MILP}

\begin{table}[t]
\centering
\footnotesize
\caption{Performance of the deterministic MILP benchmark solved using Gurobi.}
\label{tab:gurobi}
\begin{tabular}{lrrl}
\toprule
Instance & Runtime (s) & Det. NPV (\$M) & Status \\
\midrule
Newman1          & 12.74   & 24.17 & Optimal \\
Marvin           & 2009.49 & 2.11  & Time limit \\
McLaughlin Limit & --      & --    & Out of memory \\
\bottomrule
\end{tabular}
\end{table}

To provide a deterministic benchmark, we compare the proposed approach with a deterministic MILP formulation solved using Gurobi. 
Table~\ref{tab:gurobi} summarizes the performance of the deterministic MILP solved using Gurobi. The \textit{Runtime} column reports the total wall-clock solution time in seconds, \textit{Det. NPV} gives the objective value of the deterministic formulation in millions of dollars, and \textit{Status} indicates the solver termination condition. Gurobi successfully solved the Newman1 instance to optimality within the specified $1\%$ MIP gap tolerance in only 12.74 seconds, demonstrating that exact optimization is practical for small-scale instances. For the larger Marvin instance, Gurobi reached the time limit after obtaining a feasible solution, indicating that proving near-optimality becomes considerably more challenging as the problem size increases. For the McLaughlin Limit instance, the optimization terminated due to insufficient memory before the search could be completed. Therefore, no deterministic benchmark could be obtained.

These results highlight the increasing computational complexity of exact optimization for large-scale OPMSP instances. While the deterministic MILP formulation solves the small Newman1 instance efficiently, the larger Marvin instance reaches the predefined time limit and the McLaughlin Limit instance exceeds the available memory. Incorporating the chance constraints and uncertainty into the MILP formulation would substantially increase the model complexity beyond that of the deterministic formulation, making exact optimization even more challenging for large-scale instances. In contrast, the proposed evolutionary algorithms efficiently solve all benchmark instances while directly optimizing the chance-constrained formulation.

\subsubsection{Ablation Study}

\begin{table}[!t]
\centering
\footnotesize
\caption{Ablation study of the proposed initialization and mutation operators. Chance-constrained NPV (\$M) over 30 runs with matched random seeds.}
\label{tab:ablation-study}
\begin{adjustbox}{max width=\textwidth}
\begin{tabular}{lllccccccccc}
\toprule
\multirow{2}{*}{Algo}
& \multirow{2}{*}{Instance}
& \multirow{2}{*}{$\alpha$}
& \multicolumn{3}{c}{\textbf{Our Method}}
& \multicolumn{3}{c}{\textbf{Init Ablation\textsuperscript{(I)}}}
& \multicolumn{3}{c}{\textbf{Mutation Ablation\textsuperscript{(M)}}} \\
\cmidrule(lr){4-6}
\cmidrule(lr){7-9}
\cmidrule(lr){10-12}
& & & Mean & Std & Stat & Mean & Std & Stat & Mean & Std & Stat \\
\midrule
\multirow{9}{*}{\rotatebox{90}{(1+1)~EA}} & \multirow{3}{*}{Newman1} & 0.60 & \textbf{23.66} & 0.09 & I\textsuperscript{(+)}M\textsuperscript{(+)} & 22.91 & 0.21 & \textsuperscript{(-)} & 23.27 & 0.06 & \textsuperscript{(-)} \\
 &  & 0.90 & \textbf{22.99} & 0.10 & I\textsuperscript{(+)}M\textsuperscript{(+)} & 22.30 & 0.26 & \textsuperscript{(-)} & 22.61 & 0.07 & \textsuperscript{(-)} \\
 &  & 0.99 & \textbf{22.33} & 0.09 & I\textsuperscript{(+)}M\textsuperscript{(+)} & 21.63 & 0.20 & \textsuperscript{(-)} & 21.95 & 0.04 & \textsuperscript{(-)} \\
\cline{2-12}
 & \multirow{3}{*}{Marvin} & 0.60 & \textbf{811.82} & 6.64 & I\textsuperscript{(+)}M\textsuperscript{(+)} & 10.79 & 16.30 & \textsuperscript{(-)} & 807.82 & 5.90 & \textsuperscript{(-)} \\
 &  & 0.90 & \textbf{807.93} & 8.05 & I\textsuperscript{(+)}M\textsuperscript{(+)} & 13.49 & 14.40 & \textsuperscript{(-)} & 804.74 & 4.84 & \textsuperscript{(-)} \\
 &  & 0.99 & \textbf{802.37} & 5.53 & I\textsuperscript{(+)}M\textsuperscript{(*)} & 6.64 & 13.71 & \textsuperscript{(-)} & 798.12 & 6.26 & \textsuperscript{(*)} \\
\cline{2-12}
 & \multirow{3}{*}{McLaughlin L.} & 0.60 & \textbf{952.72} & 1.16 & I\textsuperscript{(+)}M\textsuperscript{(+)} & 807.48 & 18.42 & \textsuperscript{(-)} & 951.42 & 1.17 & \textsuperscript{(-)} \\
 &  & 0.90 & \textbf{950.07} & 1.43 & I\textsuperscript{(+)}M\textsuperscript{(+)} & 800.07 & 17.20 & \textsuperscript{(-)} & 948.85 & 1.06 & \textsuperscript{(-)} \\
 &  & 0.99 & \textbf{947.09} & 1.66 & I\textsuperscript{(+)}M\textsuperscript{(+)} & 803.72 & 17.80 & \textsuperscript{(-)} & 945.91 & 1.03 & \textsuperscript{(-)} \\
\midrule
\multirow{9}{*}{\rotatebox{90}{GSEMO}} & \multirow{3}{*}{Newman1} & 0.60 & \textbf{23.60} & 0.11 & I\textsuperscript{(+)}M\textsuperscript{(+)} & 22.74 & 0.34 & \textsuperscript{(-)} & 23.27 & 0.06 & \textsuperscript{(-)} \\
 &  & 0.90 & \textbf{22.93} & 0.11 & I\textsuperscript{(+)}M\textsuperscript{(+)} & 22.09 & 0.33 & \textsuperscript{(-)} & 22.60 & 0.06 & \textsuperscript{(-)} \\
 &  & 0.99 & \textbf{22.26} & 0.11 & I\textsuperscript{(+)}M\textsuperscript{(+)} & 21.44 & 0.32 & \textsuperscript{(-)} & 21.93 & 0.06 & \textsuperscript{(-)} \\
\cline{2-12}
 & \multirow{3}{*}{Marvin} & 0.60 & \textbf{810.48} & 5.14 & I\textsuperscript{(+)}M\textsuperscript{(+)} & 0.52 & 3.03 & \textsuperscript{(-)} & 808.10 & 5.00 & \textsuperscript{(-)} \\
 &  & 0.90 & \textbf{805.88} & 5.13 & I\textsuperscript{(+)}M\textsuperscript{(+)} & 0.43 & 3.01 & \textsuperscript{(-)} & 803.50 & 4.99 & \textsuperscript{(-)} \\
 &  & 0.99 & \textbf{801.22} & 5.12 & I\textsuperscript{(+)}M\textsuperscript{(+)} & 0.34 & 3.02 & \textsuperscript{(-)} & 798.84 & 4.98 & \textsuperscript{(-)} \\
\cline{2-12}
 & \multirow{3}{*}{McLaughlin L.} & 0.60 & \textbf{952.44} & 1.20 & I\textsuperscript{(+)}M\textsuperscript{(+)} & 807.43 & 16.18 & \textsuperscript{(-)} & 951.40 & 1.10 & \textsuperscript{(-)} \\
 &  & 0.90 & \textbf{949.88} & 1.20 & I\textsuperscript{(+)}M\textsuperscript{(+)} & 805.34 & 16.12 & \textsuperscript{(-)} & 948.83 & 1.10 & \textsuperscript{(-)} \\
 &  & 0.99 & \textbf{947.29} & 1.20 & I\textsuperscript{(+)}M\textsuperscript{(+)} & 803.23 & 16.06 & \textsuperscript{(-)} & 946.24 & 1.10 & \textsuperscript{(-)} \\
\midrule
\multirow{9}{*}{\rotatebox{90}{MOEA/D}} & \multirow{3}{*}{Newman1} & 0.60 & \textbf{23.74} & 0.07 & I\textsuperscript{(+)}M\textsuperscript{(+)} & 23.35 & 0.15 & \textsuperscript{(-)} & 23.34 & 0.01 & \textsuperscript{(-)} \\
 &  & 0.90 & \textbf{23.07} & 0.07 & I\textsuperscript{(+)}M\textsuperscript{(+)} & 22.69 & 0.15 & \textsuperscript{(-)} & 22.67 & 0.01 & \textsuperscript{(-)} \\
 &  & 0.99 & \textbf{22.39} & 0.07 & I\textsuperscript{(+)}M\textsuperscript{(+)} & 22.02 & 0.15 & \textsuperscript{(-)} & 22.00 & 0.01 & \textsuperscript{(-)} \\
\cline{2-12}
 & \multirow{3}{*}{Marvin} & 0.60 & \textbf{821.60} & 2.43 & I\textsuperscript{(+)}M\textsuperscript{(+)} & 9.14 & 21.01 & \textsuperscript{(-)} & 816.40 & 2.21 & \textsuperscript{(-)} \\
 &  & 0.90 & \textbf{816.97} & 2.41 & I\textsuperscript{(+)}M\textsuperscript{(+)} & 8.87 & 20.47 & \textsuperscript{(-)} & 811.76 & 2.20 & \textsuperscript{(-)} \\
 &  & 0.99 & \textbf{812.28} & 2.40 & I\textsuperscript{(+)}M\textsuperscript{(+)} & 8.59 & 19.93 & \textsuperscript{(-)} & 807.08 & 2.18 & \textsuperscript{(-)} \\
\cline{2-12}
 & \multirow{3}{*}{McLaughlin L.} & 0.60 & \textbf{954.93} & 0.56 & I\textsuperscript{(+)}M\textsuperscript{(+)} & 843.80 & 12.30 & \textsuperscript{(-)} & 953.18 & 0.45 & \textsuperscript{(-)} \\
 &  & 0.90 & \textbf{952.36} & 0.56 & I\textsuperscript{(+)}M\textsuperscript{(+)} & 841.55 & 12.25 & \textsuperscript{(-)} & 950.61 & 0.44 & \textsuperscript{(-)} \\
 &  & 0.99 & \textbf{949.77} & 0.56 & I\textsuperscript{(+)}M\textsuperscript{(+)} & 839.29 & 12.20 & \textsuperscript{(-)} & 948.02 & 0.44 & \textsuperscript{(-)} \\
\midrule
\multirow{9}{*}{\rotatebox{90}{NSGA-II}} & \multirow{3}{*}{Newman1} & 0.60 & \textbf{23.34} & 0.01 & I\textsuperscript{(+)}M\textsuperscript{(*)} & 22.88 & 0.12 & \textsuperscript{(-)} & \textbf{23.34} & 0.01 & \textsuperscript{(*)} \\
 &  & 0.90 & \textbf{22.67} & 0.01 & I\textsuperscript{(+)}M\textsuperscript{(*)} & 22.23 & 0.12 & \textsuperscript{(-)} & \textbf{22.67} & 0.01 & \textsuperscript{(*)} \\
 &  & 0.99 & \textbf{22.00} & 0.01 & I\textsuperscript{(+)}M\textsuperscript{(*)} & 21.57 & 0.11 & \textsuperscript{(-)} & \textbf{22.00} & 0.01 & \textsuperscript{(*)} \\
\cline{2-12}
 & \multirow{3}{*}{Marvin} & 0.60 & \textbf{816.40} & 2.21 & I\textsuperscript{(+)}M\textsuperscript{(*)} & 1.12 & 12.32 & \textsuperscript{(-)} & \textbf{816.40} & 2.21 & \textsuperscript{(*)} \\
 &  & 0.90 & \textbf{811.76} & 2.20 & I\textsuperscript{(+)}M\textsuperscript{(*)} & 0.82 & 12.14 & \textsuperscript{(-)} & \textbf{811.76} & 2.20 & \textsuperscript{(*)} \\
 &  & 0.99 & \textbf{807.08} & 2.18 & I\textsuperscript{(+)}M\textsuperscript{(*)} & 0.53 & 11.99 & \textsuperscript{(-)} & \textbf{807.08} & 2.18 & \textsuperscript{(*)} \\
\cline{2-12}
 & \multirow{3}{*}{McLaughlin L.} & 0.60 & \textbf{953.20} & 0.44 & I\textsuperscript{(+)}M\textsuperscript{(*)} & 841.66 & 12.22 & \textsuperscript{(-)} & \textbf{953.20} & 0.44 & \textsuperscript{(*)} \\
 &  & 0.90 & \textbf{950.63} & 0.44 & I\textsuperscript{(+)}M\textsuperscript{(*)} & 839.42 & 12.18 & \textsuperscript{(-)} & \textbf{950.63} & 0.44 & \textsuperscript{(*)} \\
 &  & 0.99 & \textbf{948.04} & 0.44 & I\textsuperscript{(+)}M\textsuperscript{(*)} & 837.15 & 12.13 & \textsuperscript{(-)} & \textbf{948.04} & 0.44 & \textsuperscript{(*)} \\
\bottomrule
\end{tabular}
\end{adjustbox}
\end{table}

To evaluate the contribution of the proposed initialization and mutation operators, we perform an ablation study. Table~\ref{tab:ablation-study} summarizes the results obtained using the proposed method together with the \textit{Init Ablation} and \textit{Mutation Ablation} variants, which remove the proposed initialization and mutation operators, respectively. The \textit{Mean} and \textit{Std} columns report the mean and standard deviation of the chance-constrained NPV over 30 independent runs for each algorithm, benchmark instance, and confidence level $\alpha$, while the \textit{Stat} column summarizes the statistical comparison between the proposed method and the corresponding ablated variant.

The \textit{Init Ablation} consistently degrades the chance-constrained NPV for all algorithms and benchmark instances. The performance loss is relatively small for Newman1 but becomes substantially larger for Marvin and McLaughlin Limit, indicating that the proposed initialization becomes increasingly important as the problem size grows.

The \textit{Mutation Ablation} also reduces the chance-constrained NPV for (1+1) EA, GSEMO, and MOEA/D, whereas no statistically significant difference is observed for NSGA-II. Overall, both operators improve the performance of the proposed framework, with the proposed initialization having the greatest impact and the proposed mutation providing additional improvements.

Overall, the results indicate that the proposed bi-objective formulation effectively captures the trade-off between expected discounted NPV and scheduling risk. Across most benchmark instances, MOEA/D provides a strong balance between profitability, risk, and Pareto-front quality, while NSGA-II tends to generate lower-risk solutions.

\section{Conclusion}
\label{sec:conclusion}

This paper presented a bi-objective chance-constrained formulation for the large-scale open-pit mine scheduling problem under geological uncertainty. The proposed formulation simultaneously maximizes the expected discounted net present value and minimizes scheduling risk measured by the discounted standard deviation of NPV. In contrast to traditional single-objective chance-constrained approaches that require a predefined confidence level $\alpha$, the proposed formulation eliminates the need to specify $\alpha$ during optimization and instead generates a Pareto front representing different trade-offs between profitability and risk within a single optimization run.

The proposed approach was evaluated on large-scale MineLib benchmark instances containing up to 112,687 blocks using three multi-objective evolutionary algorithms, namely GSEMO, MOEA/D, and NSGA-II, together with problem-specific initialization and mutation operators. The algorithms were compared with a single-objective $(1+1)$~EA baseline. The experimental results demonstrated that the bi-objective formulation effectively captures the trade-off between profitability and risk under geological uncertainty. Among the considered algorithms, MOEA/D consistently produced high-quality Pareto-front approximations and achieved the highest chance-constrained NPV values across most benchmark instances and confidence levels, while NSGA-II tended to generate lower-risk solutions.

The temporal analyses of discounted expected NPV, scheduling risk, and production tonnage further showed that the proposed formulation generates more balanced extraction schedules than the single-objective baseline. Comparison with a deterministic MILP benchmark demonstrated that exact optimization becomes computationally challenging for large-scale instances, whereas the proposed evolutionary approach efficiently solved all benchmark instances while directly optimizing the chance-constrained formulation. In addition, the performance indicator results confirm that the proposed framework generates high-quality Pareto-front approximations in terms of convergence and diversity. Overall, these results demonstrate the effectiveness of bi-objective evolutionary optimization for large-scale chance-constrained open-pit mine scheduling under geological uncertainty.

Future work will focus on extending the framework to diverse operational scenarios, investigating alternative risk measures to enhance scheduling robustness, and incorporating more realistic geological uncertainty models that explicitly capture spatial correlations, orebody continuity, and non-Gaussian grade distributions.

\bibliographystyle{unsrtnat} 
\bibliography{lib}

\appendix
\section{Statistical Tests and Effect Size}
\label{app:effect_size}

Statistical significance is assessed using the paired Wilcoxon signed-rank test. To account for multiple comparisons, Holm's procedure is used to adjust the resulting $p$-values.

The effect size is measured using the matched-pairs rank-biserial correlation (RBC)~\cite{Kerby2014}, defined as
\[
r_{\mathrm{rb}}
=
\frac{W^+-W^-}{W^++W^-},
\]
\noindent where $W^+$ and $W^-$ denote the sums of the positive and negative signed ranks, respectively.

Table~\ref{tab:wilcoxon-holm-effect_cc_1} reports the pairwise statistical comparisons and effect sizes for the chance-constrained NPV values for the
different confidence levels, while Table~\ref{tab:indicator-wilcoxon-holm_pe_1} reports the corresponding results for the performance indicators.

\begin{table*}[!t]
\centering
\footnotesize
\caption{Paired Wilcoxon signed-rank test results with Holm-adjusted $p$-values and rank-biserial correlation effect sizes for chance constrained NPV.}
\label{tab:wilcoxon-holm-effect_cc_1}
\begin{adjustbox}{max width=\textwidth}
\begin{tabular}{lllrllll}
\toprule
Instance & $\alpha$ & Comparison & $W$ & Raw $p$ & Holm $p$ & Significant & Effect size \\
\midrule
Newman1 & 0.60 & (1+1)~EA vs. GSEMO & 137.0 & 0.0497 & \textbf{0.0497} & Yes & 0.411 \\
 &  & (1+1)~EA vs. MOEA/D & 72.0 & 0.000555 & \textbf{0.00111} & Yes & -0.690 \\
 &  & (1+1)~EA vs. NSGA-II & 0.0 & 1.86e-09 & \textbf{1.12e-08} & Yes & 1.000 \\
 &  & GSEMO vs. MOEA/D & 15.0 & 2.55e-07 & \textbf{7.66e-07} & Yes & -0.935 \\
 &  & GSEMO vs. NSGA-II & 0.0 & 1.86e-09 & \textbf{1.12e-08} & Yes & 1.000 \\
 &  & MOEA/D vs. NSGA-II & 0.0 & 1.86e-09 & \textbf{1.12e-08} & Yes & 1.000 \\
\addlinespace
 & 0.90 & (1+1)~EA vs. GSEMO & 149.0 & 0.0879 & 0.0879 & No & 0.359 \\
 &  & (1+1)~EA vs. MOEA/D & 97.0 & 0.00434 & \textbf{0.00868} & Yes & -0.583 \\
 &  & (1+1)~EA vs. NSGA-II & 0.0 & 1.86e-09 & \textbf{1.12e-08} & Yes & 1.000 \\
 &  & GSEMO vs. MOEA/D & 15.0 & 2.55e-07 & \textbf{7.66e-07} & Yes & -0.935 \\
 &  & GSEMO vs. NSGA-II & 0.0 & 1.86e-09 & \textbf{1.12e-08} & Yes & 1.000 \\
 &  & MOEA/D vs. NSGA-II & 0.0 & 1.86e-09 & \textbf{1.12e-08} & Yes & 1.000 \\
\addlinespace
 & 0.99 & (1+1)~EA vs. GSEMO & 119.0 & 0.0185 & \textbf{0.0212} & Yes & 0.488 \\
 &  & (1+1)~EA vs. MOEA/D & 110.0 & 0.0106 & \textbf{0.0212} & Yes & -0.527 \\
 &  & (1+1)~EA vs. NSGA-II & 0.0 & 1.86e-09 & \textbf{1.12e-08} & Yes & 1.000 \\
 &  & GSEMO vs. MOEA/D & 16.0 & 3.15e-07 & \textbf{9.44e-07} & Yes & -0.931 \\
 &  & GSEMO vs. NSGA-II & 0.0 & 1.86e-09 & \textbf{1.12e-08} & Yes & 1.000 \\
 &  & MOEA/D vs. NSGA-II & 0.0 & 1.86e-09 & \textbf{1.12e-08} & Yes & 1.000 \\
\addlinespace
\midrule
Marvin & 0.60 & (1+1)~EA vs. GSEMO & 171.0 & 0.213 & 0.213 & No & 0.265 \\
 &  & (1+1)~EA vs. MOEA/D & 0.0 & 1.86e-09 & \textbf{1.12e-08} & Yes & -1.000 \\
 &  & (1+1)~EA vs. NSGA-II & 81.0 & 0.00123 & \textbf{0.00246} & Yes & -0.652 \\
 &  & GSEMO vs. MOEA/D & 1.0 & 3.73e-09 & \textbf{1.49e-08} & Yes & -0.996 \\
 &  & GSEMO vs. NSGA-II & 26.0 & 1.99e-06 & \textbf{5.97e-06} & Yes & -0.888 \\
 &  & MOEA/D vs. NSGA-II & 0.0 & 1.86e-09 & \textbf{1.12e-08} & Yes & 1.000 \\
\addlinespace
 & 0.90 & (1+1)~EA vs. GSEMO & 138.0 & 0.0523 & 0.0523 & No & 0.406 \\
 &  & (1+1)~EA vs. MOEA/D & 7.0 & 3.54e-08 & \textbf{1.42e-07} & Yes & -0.970 \\
 &  & (1+1)~EA vs. NSGA-II & 121.0 & 0.0208 & \textbf{0.0417} & Yes & -0.480 \\
 &  & GSEMO vs. MOEA/D & 1.0 & 3.73e-09 & \textbf{1.86e-08} & Yes & -0.996 \\
 &  & GSEMO vs. NSGA-II & 24.0 & 1.42e-06 & \textbf{4.26e-06} & Yes & -0.897 \\
 &  & MOEA/D vs. NSGA-II & 0.0 & 1.86e-09 & \textbf{1.12e-08} & Yes & 1.000 \\
\addlinespace
 & 0.99 & (1+1)~EA vs. GSEMO & 212.0 & 0.685 & 0.685 & No & 0.088 \\
 &  & (1+1)~EA vs. MOEA/D & 5.0 & 1.86e-08 & \textbf{7.45e-08} & Yes & -0.978 \\
 &  & (1+1)~EA vs. NSGA-II & 52.0 & 7.06e-05 & \textbf{0.000141} & Yes & -0.776 \\
 &  & GSEMO vs. MOEA/D & 1.0 & 3.73e-09 & \textbf{1.86e-08} & Yes & -0.996 \\
 &  & GSEMO vs. NSGA-II & 24.0 & 1.42e-06 & \textbf{4.26e-06} & Yes & -0.897 \\
 &  & MOEA/D vs. NSGA-II & 0.0 & 1.86e-09 & \textbf{1.12e-08} & Yes & 1.000 \\
\addlinespace
\midrule
McLaughlin Limit  & 0.60 & (1+1)~EA vs. GSEMO & 194.0 & 0.44 & 0.44 & No & 0.166 \\
 &  & (1+1)~EA vs. MOEA/D & 3.0 & 9.31e-09 & \textbf{4.66e-08} & Yes & -0.987 \\
 &  & (1+1)~EA vs. NSGA-II & 144.0 & 0.0699 & 0.14 & No & -0.381 \\
 &  & GSEMO vs. MOEA/D & 5.0 & 1.86e-08 & \textbf{7.45e-08} & Yes & -0.978 \\
 &  & GSEMO vs. NSGA-II & 88.0 & 0.00219 & \textbf{0.00656} & Yes & -0.622 \\
 &  & MOEA/D vs. NSGA-II & 0.0 & 1.86e-09 & \textbf{1.12e-08} & Yes & 1.000 \\
\addlinespace
 & 0.90 & (1+1)~EA vs. GSEMO & 212.0 & 0.685 & 0.685 & No & 0.088 \\
 &  & (1+1)~EA vs. MOEA/D & 13.0 & 1.64e-07 & \textbf{6.56e-07} & Yes & -0.944 \\
 &  & (1+1)~EA vs. NSGA-II & 137.0 & 0.0497 & 0.0994 & No & -0.411 \\
 &  & GSEMO vs. MOEA/D & 5.0 & 1.86e-08 & \textbf{9.31e-08} & Yes & -0.978 \\
 &  & GSEMO vs. NSGA-II & 90.0 & 0.00256 & \textbf{0.00768} & Yes & -0.613 \\
 &  & MOEA/D vs. NSGA-II & 0.0 & 1.86e-09 & \textbf{1.12e-08} & Yes & 1.000 \\
\addlinespace
 & 0.99 & (1+1)~EA vs. GSEMO & 200.0 & 0.516 & 0.516 & No & -0.140 \\
 &  & (1+1)~EA vs. MOEA/D & 3.0 & 9.31e-09 & \textbf{4.66e-08} & Yes & -0.987 \\
 &  & (1+1)~EA vs. NSGA-II & 105.0 & 0.00761 & \textbf{0.0152} & Yes & -0.548 \\
 &  & GSEMO vs. MOEA/D & 5.0 & 1.86e-08 & \textbf{7.45e-08} & Yes & -0.978 \\
 &  & GSEMO vs. NSGA-II & 90.0 & 0.00256 & \textbf{0.00768} & Yes & -0.613 \\
 &  & MOEA/D vs. NSGA-II & 0.0 & 1.86e-09 & \textbf{1.12e-08} & Yes & 1.000 \\
\bottomrule
\end{tabular}
\end{adjustbox}
\end{table*}

\begin{table*}[!t]
\centering
\caption{Paired Wilcoxon signed-rank test results for the performance indicators. Holm-adjusted $p$-values and matched-pairs rank-biserial correlations are reported.}
\label{tab:indicator-wilcoxon-holm_pe_1}
\begin{adjustbox}{max width=\textwidth}
\begin{tabular}{llrrrrrr}
\toprule
Instance & $\alpha$ & Comparison & $W$ & Raw $p$ & Holm $p$ & Significant & Effect size \\
\midrule
Newman1 & HV & GSEMO vs MOEA/D & 176.0 & 0.253 & 0.253 & No & -0.243 \\
 &  & GSEMO vs NSGA-II & 0.0 & 1.86e-09 & \textbf{5.59e-09} & Yes & 1.000 \\
 &  & MOEA/D vs NSGA-II & 0.0 & 1.86e-09 & \textbf{5.59e-09} & Yes & 1.000 \\
\addlinespace
 & IGD & GSEMO vs MOEA/D & 0.0 & 1.86e-09 & \textbf{5.59e-09} & Yes & -1.000 \\
 &  & GSEMO vs NSGA-II & 0.0 & 1.86e-09 & \textbf{5.59e-09} & Yes & -1.000 \\
 &  & MOEA/D vs NSGA-II & 0.0 & 1.86e-09 & \textbf{5.59e-09} & Yes & -1.000 \\
\addlinespace
 & GD & GSEMO vs MOEA/D & 27.0 & 2.35e-06 & \textbf{2.35e-06} & Yes & 0.884 \\
 &  & GSEMO vs NSGA-II & 3.0 & 9.31e-09 & \textbf{1.86e-08} & Yes & -0.987 \\
 &  & MOEA/D vs NSGA-II & 0.0 & 1.86e-09 & \textbf{5.59e-09} & Yes & -1.000 \\
\addlinespace
 & $\epsilon$ & GSEMO vs MOEA/D & 76.0 & 0.000798 & \textbf{0.000798} & Yes & 0.673 \\
 &  & GSEMO vs NSGA-II & 0.0 & 1.86e-09 & \textbf{5.59e-09} & Yes & -1.000 \\
 &  & MOEA/D vs NSGA-II & 0.0 & 1.86e-09 & \textbf{5.59e-09} & Yes & -1.000 \\
\addlinespace
\midrule
Marvin & HV & GSEMO vs MOEA/D & 37.0 & 1.06e-05 & \textbf{1.06e-05} & Yes & -0.841 \\
 &  & GSEMO vs NSGA-II & 0.0 & 1.86e-09 & \textbf{5.59e-09} & Yes & -1.000 \\
 &  & MOEA/D vs NSGA-II & 3.0 & 9.31e-09 & \textbf{1.86e-08} & Yes & -0.987 \\
\addlinespace
 & IGD & GSEMO vs MOEA/D & 98.0 & 0.00466 & \textbf{0.00466} & Yes & 0.578 \\
 &  & GSEMO vs NSGA-II & 0.0 & 1.86e-09 & \textbf{5.59e-09} & Yes & 1.000 \\
 &  & MOEA/D vs NSGA-II & 0.0 & 1.86e-09 & \textbf{5.59e-09} & Yes & 1.000 \\
\addlinespace
 & GD & GSEMO vs MOEA/D & 79.0 & 0.00104 & \textbf{0.00208} & Yes & 0.660 \\
 &  & GSEMO vs NSGA-II & 137.0 & 0.0497 & \textbf{0.0497} & Yes & 0.411 \\
 &  & MOEA/D vs NSGA-II & 58.0 & 0.000137 & \textbf{0.000412} & Yes & -0.751 \\
\addlinespace
 & $\epsilon$ & GSEMO vs MOEA/D & 66.0 & 0.000313 & \textbf{0.000313} & Yes & 0.716 \\
 &  & GSEMO vs NSGA-II & 2.0 & 5.59e-09 & \textbf{1.12e-08} & Yes & 0.991 \\
 &  & MOEA/D vs NSGA-II & 1.0 & 3.73e-09 & \textbf{1.12e-08} & Yes & 0.996 \\
\midrule
McLaughlin Limit & HV & GSEMO vs MOEA/D & 149.0 & 0.0879 & 0.176 & No & -0.359 \\
 &  & GSEMO vs NSGA-II & 212.0 & 0.685 & 0.685 & No & 0.088 \\
 &  & MOEA/D vs NSGA-II & 124.0 & 0.0248 & 0.0744 & No & 0.467 \\
\addlinespace
 & IGD & GSEMO vs MOEA/D & 218.0 & 0.777 & 0.777 & No & -0.062 \\
 &  & GSEMO vs NSGA-II & 185.0 & 0.339 & 0.677 & No & 0.204 \\
 &  & MOEA/D vs NSGA-II & 143.0 & 0.0667 & 0.2 & No & 0.385 \\
\addlinespace
 & GD & GSEMO vs MOEA/D & 139.0 & 0.0549 & 0.0549 & No & 0.402 \\
 &  & GSEMO vs NSGA-II & 35.0 & 7.99e-06 & \textbf{1.6e-05} & Yes & -0.849 \\
 &  & MOEA/D vs NSGA-II & 1.0 & 3.73e-09 & \textbf{1.12e-08} & Yes & -0.996 \\
\addlinespace
 & $\epsilon$ & GSEMO vs MOEA/D & 113.0 & 0.0128 & \textbf{0.0128} & Yes & 0.514 \\
 &  & GSEMO vs NSGA-II & 40.0 & 1.6e-05 & \textbf{4.79e-05} & Yes & 0.828 \\
 &  & MOEA/D vs NSGA-II & 91.0 & 0.00277 & \textbf{0.00553} & Yes & 0.609 \\
\addlinespace
\bottomrule
\end{tabular}
\end{adjustbox}
\end{table*}

\end{document}